\documentclass[lettersize,journal]{IEEEtran}
\usepackage{amsmath,amsfonts}
\usepackage{algorithmic}
\usepackage{algorithm}
\usepackage{array}
\usepackage[caption=false,font=normalsize,labelfont=sf,textfont=sf]{subfig}
\usepackage{textcomp}
\usepackage{stfloats}
\usepackage{url}
\usepackage{verbatim}
\usepackage{graphicx}
\usepackage{booktabs}
\usepackage{multirow}
\usepackage{arydshln}
\usepackage{float}
\usepackage{hyperref}
\usepackage{xcolor} 
\usepackage{color,soul}
\usepackage{cite}
\usepackage{color, colortbl}

\definecolor{Orange}{rgb}{1.0, 0.27, 0.0}
\NewDocumentCommand{\eskandarian}{ mO{}}{\textcolor{Orange}{\textsuperscript{\textit{eskandarian}}\textsf{\textbf{\small[#1]}}}}
\NewDocumentCommand{\ce}{ mO{}}{\textcolor{Orange}{\textsuperscript{\textit{eskandarian}}\textsf{\textbf{\small[#1]}}}}
\newcolumntype{M}{>{\centering\arraybackslash}m{1.2cm}}
\newcolumntype{A}{>{\centering\arraybackslash}m{1.8cm}}
\newcolumntype{B}{>{\arraybackslash}m{3cm}}
\begin{document}

\title{\huge A Quality Index Metric and Method for Online Self-Assessment of Autonomous Vehicles Sensory Perception}

\author{Ce Zhang,~\IEEEmembership{IEEE Student Member}, Azim Eskandarian,~\IEEEmembership{IEEE Senior Member},
\thanks{
Ce Zhang is with Department of Mechanical Engineering, ASIM Laboratory Virginia Tech, 635 Prices Fork Rd. Blacksburg, VA 24060, USA {\tt\small zce@vt.edu}
}
\thanks{
Azim Eskandarian is with the Department of Mechanical Engineering, Virginia Tech, He is the department head and the director of the ASIM Laboratory, Blacksburg, VA 2460, USA USA {\tt\small eskandarian@vt.edu}
}
}

\markboth{}%
{Shell \MakeLowercase{\textit{et al.}}: A Sample Article Using IEEEtran.cls for IEEE Journals}


\maketitle
\begin{abstract}
Reliable object detection using cameras plays a crucial role in enabling autonomous vehicles to perceive their surroundings. However, existing camera-based object detection approaches for autonomous driving lack the ability to provide comprehensive feedback on detection performance for individual frames. To address this limitation, we propose a novel evaluation metric, named as the detection quality index (DQI), which assesses the performance of camera-based object detection algorithms and provides frame-by-frame feedback on detection quality. The DQI is generated by combining the intensity of the fine-grained saliency map with the output results of the object detection algorithm. Additionally, we have developed a superpixel-based attention network (SPA-NET) that utilizes raw image pixels and superpixels as input to predict the proposed DQI evaluation metric. To validate our approach, we conducted experiments on three open-source datasets. The results demonstrate that the proposed evaluation metric accurately assesses the detection quality of camera-based systems in autonomous driving environments. Furthermore, the proposed SPA-NET outperforms other popular image-based quality regression models. This highlights the effectiveness of the DQI in evaluating a camera's ability to perceive visual scenes. Overall, our work introduces a valuable self-evaluation tool for camera-based object detection in autonomous vehicles.
\end{abstract}

\begin{IEEEkeywords}
Autonomous Vehicle, Neural Network, Computer Vision, Image Processing, Image Quality Assessment  
\end{IEEEkeywords}
\section{Introduction}\label{chpt:introduction}
\IEEEPARstart{A}{utonomous} vehicle development has grown rapidly in recent years because of increasing computing power and more robust decision-making algorithms \cite{9242336, 9780370}. 
Perception is a fundamental but vital step for autonomous driving \cite{9564709, VANBRUMMELEN2018384}. 
Cameras have emerged as the preferred perception sensors for autonomous vehicles because of the fast processing speed and low cost. 
In autonomous driving applications, cameras are predominantly utilized for tasks such as object detection, lane detection, and segmentation \cite{8936542, 10.1007/978-3-031-25066-8_41, Ding2023SalienDetAS, 10.1115/IMECE2021-69975}.

\begin{figure*}[h]
\centering
\includegraphics[width=13.5cm]{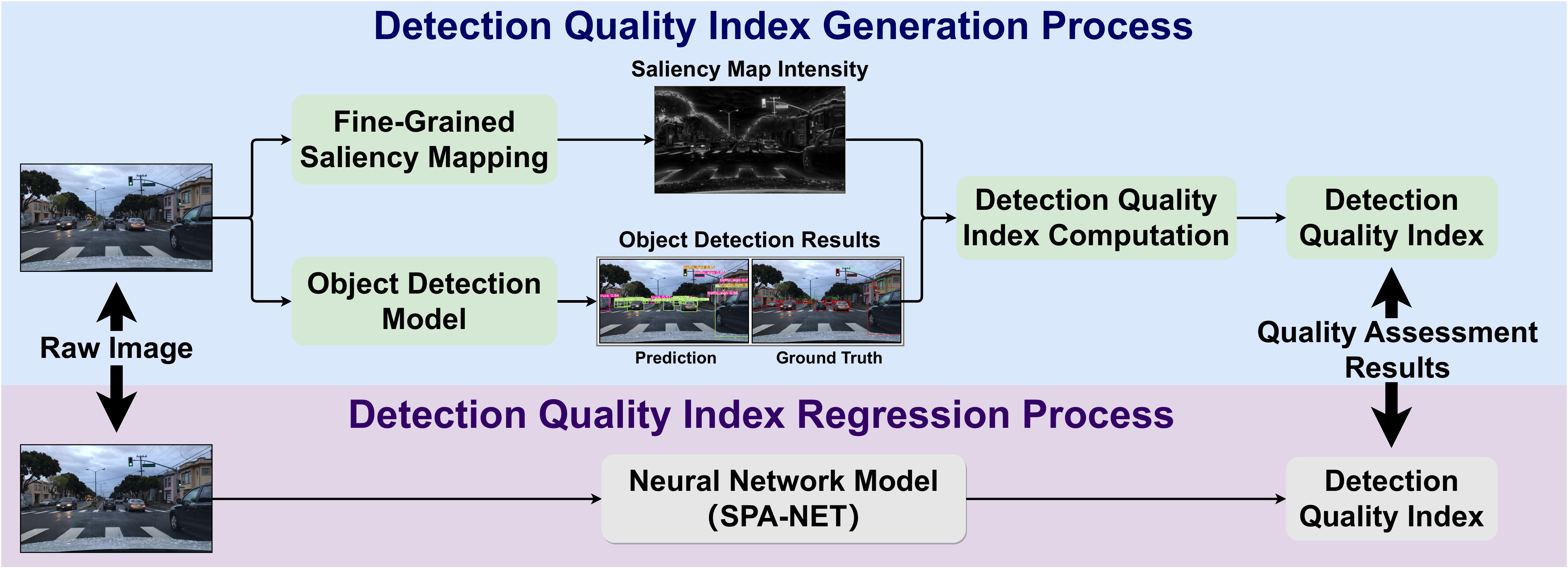}
\caption{Detection Quality Index Generation and Prediction Flowchart. Since the detection quality index generation process comprise ground truth annotations, it is necessary to design a neural network model to predict it.}
\label{fig1:flowchart}
\end{figure*}

Object detection plays a crucial role for autonomous vehicle perception. 
Traditional camera-based detection algorithms employ explicit appearance features such as edges to locate and classify objects \cite{8627998}. 
Modern detection algorithms leverage deep neural networks and parallel computing methods to extract hidden and implicit features, results in significant improvement in detection accuracy. 

Despite these advancement, most existing camera-based detection algorithms do not provide a comprehensive feedback on the detection quality of their outputs for each individual image. 
In other words, there is a lack of knowledge regarding the reliability and accuracy of the detection algorithm when applied to specific images. 
This becomes a crucial concern in the context of autonomous driving, as a false negative detection or incorrect classification can potentially lead to dangerous vehicle maneuvers and, consequently, accidents. 
Therefore, it is imperative to explore methods that can offer detection quality feedback for object detection algorithms during autonomous driving scenarios.
This feedback mechanism would enable a better understanding of the detection algorithm's reliability, allowing for improved decision-making and overall system safety.

In the field of computer vision, blind image quality assessment (B-IQA) methods have been widely used to evaluate the overall quality of images \cite{1038057}. 
However, directly applying these methods to provide perceptual quality feedback for autonomous vehicle systems poses certain challenges: 
(a) the B-IQA score is evaluated by human subjects' mean opinion scores instead of object detection algorithms, 
and (b) the B-IQA score is highly affected by image shape such as stretching and rotation, which is usually applied as a data augmentation method for object detection algorithms training \cite{Wang2021ASO, 9157018}. 
Nonetheless, the B-IQA approaches raise important questions in the context of autonomous vehicle perception: 
(a) What impacts the object detection algorithm's perceptual quality under an autonomous driving environment? 
(b) Can we quantitatively define an index, namely DQI, similar to the B-IQA approach for the camera-based detection algorithm perceptual quality with given images? 
(c) If the DQI is defined, can we design a computational model to estimate the quantitative index for the camera perceptual quality?

To address these questions, we propose a novel evaluation metric to define camera-based perceptual quality for object detection in autonomous driving. 
Additionally, we develop an algorithm to predict the perceptual quality, as seen in Figure \ref{fig1:flowchart}. To begin, we utilize two object detection algorithms to investigate the detection performance in an autonomous driving environment.
Then, we meticulously define a camera-based autonomous vehicle detection quality evaluation metric based on the image saliency map and object detection algorithms' results. 
Finally, we propose a deep neural network model to predict perceptual quality results. The contributions of our paper are:

\begin{itemize}
    \item Clearly describe the impact factors for camera-based object detection algorithms under an autonomous driving environment.
    \item Quantitatively define the camera-based object detection algorithm perceptual quality for a single image frame based on saliency mapping and object detection algorithm's results.
    \item Develop a superpixel-based attention neural network model to predict the perceptual quality index.
\end{itemize}

\section{Related Works}
Since the proposed DQI is inspired by B-IQA and the construction of DQI comprises object detection algorithms, this section reviews computational regression models for B-IQA tasks and camera-based object detection algorithms.

\subsection{Blind Image Quality Assessment}
B-IQA evaluates images' visual quality without reference. 
Traditional B-IQA methods manually extract image features and then apply a classifier or a regression model to assess the image quality. 
Modern B-IQA methods utilize deep neural networks to extract features and classify or regress the image quality automatically.

\subsubsection{Traditional Blind Image Quality Assessment Algorithms}
Traditional B-IQA methods employ natural scene features to rate images' visual quality without reference. 
The original idea of B-IQA is proposed by Li in 2002 \cite{1038057}. 
He mentioned that edge sharpness, random noise level, and structure noises are the key to B-IQA.
The following research on the B-IQA is focused on improving the regression accuracy and decrease the computation load.
A. Moorthy proposed a B-IQA
method for distorted images, namely the DIIVINE algorithm \cite{5756237}. 
They decomposed the distorted images through wavelet transforms, then selected subband coefficients to extract statistical features for distorted image quality evaluation.
Even though high regression correlation, DIIVINE's computation speed is slow due to constructing complex 2D probability density functions. 
A. Mittal et al. proposed a blind image quality evaluator, namely BRISQUE, to assess the image quality without reference through a spatial natural scene statistic model \cite{6272356}. 
The benefit of this algorithm is computationally efficient because it does not require any coordinate transformation such as the Fourier transform or wavelet transform. 
\subsubsection{Neural Network-based Algorithms} 
With the advent of parallel computing tools, deep neural networks have emerged as effective solutions for both classification and regression tasks in B-IQA. 
Early studies in B-IQA primarily focused on architectural innovations and end-to-end regression techniques.
One notable work by W. Hou et al. propose a deep learning-based B-IQA algorithm \cite{6872541}. 
In \cite{6872541}, they apply selective natural scene statistics as features and design a multi-layer perceptron (MLP) model to classify the image quality discriminatively. 
W. Zhang et al. improve the B-IQA from manual feature extraction to an end-to-end model. 
They propose a bilinear convolutional neural network (CNN) to predict images' quality scores using images as inputs \cite{8576582}.
Besides CNN, the transformer architecture has also been applied for B-IQA due to the benefits of global feature extraction and attention-based mechanism.
J. You et al. combines transformer with CNN for B-IQA estimation \cite{9506075}. 
They apply a pre-trained ResNet as the backbone for image feature extraction at first. 
Then, a visual transformer's (ViT) architecture is used to extract the ResNet output features. 
J. Ke directly utilize the ViT module as the backbone for B-IQA \cite{Ke2021MUSIQMI}. 
They keep the image aspect ratio and used multi-scale images as the input. 
The key innovation of their model is the scale positional encoder design to preserve the scale features. 
In recent studies, researchers have shifted their focus from solely model innovation to exploring effective learning strategies and meta-learning approaches.
H. Zhu et al. propose a meta-learning mechanism for B-IQA \cite{Zhu2020MetaIQADM}. 
In their study, different image distortions such as brighten and white noise are used as the supporting dataset. 
Then, the unknown distorted images are fed into the prior knowledge model to predict the B-IQA results. 
As for learning strategies, Z. Wang et al. \cite{Wang2021SemiSupervisedDE} develop a semi-supervised learning approach by combing labeled and unlabeled data to train an ensemble model. 
Z. Zhou \cite{Zhou2023CollaborativeAF} propose an novel auto-encoder model for B-IQA with self-supervised learning approach.

In summary, computational models for the B-IQA task have evolved from manual feature extraction to end-to-end neural network models. 
Current research efforts are focused on improving computation speed and model generalization through various learning strategies and model designs.

\subsection{Object Detection Algorithms}
Object detection algorithms are key to autonomous vehicle perception. 
Most 2D object detection algorithms can be categorized into one-stage and two-stage detection methods.
One-stage algorithms predict the objects' classes and locations simultaneously while two-stage algorithms generate a region proposal map at first, then predict the objects' classes and locations \cite{Fu2017DSSDD, 10.1007/978-3-319-46448-0_2, Zhou2021ProbabilisticTD, Girshick2014RichFH, Girshick2015FastR, Ren2015FasterRT}. 
The benefits of two-stage detectors is the accurate detection performance, as the region proposal module can propose numerous potential objects.
However, a drawback of two-stage detectors is the computing speed due to redundant object proposals.
As for single-stage detectors, the computation load is significantly reduced.
However, the trade-off for improved speed is a potential decrease in detection performance, compared to two-stage detectors.
Previous literatures have extensively explained and reviewed these methods. 
This section focuses on the YOLO series object detection algorithms since it is widely applied in robotics and autonomous vehicle perception.

The YOLO series algorithms are real-time one-stage object detection algorithms. 
The backbone of YOLO series algorithms is the DarkNet model \cite{redmon2013darknet}, which is composed of a series of convolutional layers combined with residual modules. 
Development of the YOLO series (from v1-v7) algorithms can be summarized as improving the inference speed and enhance the detection performance by capturing small-size or partial occluded objects.

The first generation of the YOLO algorithm, namely YOLO-v1, was developed by J. Redmon et al. \cite{Redmon2016YouOL}. 
The model comprises 24 convolutional layers followed by two fully connected layers for object classification and localization. 
Since there is no complex regional proposal module, the YOLO-v1's processing speed is faster than the region-based CNN (R-CNN) series model, while this architecture is poor at localizing small and crowded objects. 
J. Redmon also proposed YOLO-v2 to improve localization accuracy. 
Instead of using fully connected layers for direct object localization and classification, they proposed prior defined anchor boxes \cite{Redmon2017YOLO9000BF}. 
Even though the anchor boxes method decreases the overall mean average precision (mAP), they predicted more bounding boxes, which has improved their model. 
The YOLO-v3 model is a major improvement on the YOLO series \cite{Redmon2018YOLOv3AI} that employ a new DarkNet-53 backbone with better feature extraction capability. 
Furthermore, to overcome the small object detection error, they propose feature pyramid networks (FPN) to extract image features from different scales. The YOLO-v4 model is proposed by A. Bochkovskiy et al. \cite{Bochkovskiy2020YOLOv4OS}. 
The main differences between the YOLO-v4 and v3 are the backbone and neck module. 
In the YOLO-v4 backbone, the cross-stage partial connections (CSP) method has been implemented to decrease the parameters' size. 
In the neck module, they applied the spatial pyramid pooling (SPP) method to extract image features from different scales effectively. 
YOLO-v5 is a marginal improvement compared with YOLO-v4. 
According to their results, the YOLO-v5 achieved faster processing speed than the YOLO-v4.
YOLO-v6 \cite{Li2022YOLOv6AS} is focused on keep shrinking model sizes by replace the CSP module to the EfficientNet and YOLO-v7 \cite{wang2023yolov7} is focusing on easier deployment for industrial applications.

Besides the official YOLO-v1 to v7, there are also some other modifications to implement the YOLO series algorithms for embedded systems. The YOLOP is a panoptic segmentation algorithm that achieves real-time on the Jetson TX2 robotic platform \cite{Wu2021YOLOPYO}. YOLOX is an anchor-free object detection algorithm \cite{Ge2021YOLOXEY}. Besides free anchor, the YOLOX model proposed a decoupled head and label assignment to achieve the state-of-the-art (SOTA) results on the COCO dataset.
\section{Proposed Methodology}
This section comprises the DQI generation methodlogy and a detailed explanation of the proposed SPA-NET regression model.
The DQI aims to produce a comprehensive quality score for assessing object detection algorithms' performances.
Meanwhile, the objective of the SPA-NET is to estimate the DQI score without relying on prior knowlege of ground-truth.
\begin{figure*}[h]
\centering
\includegraphics[width=16cm]{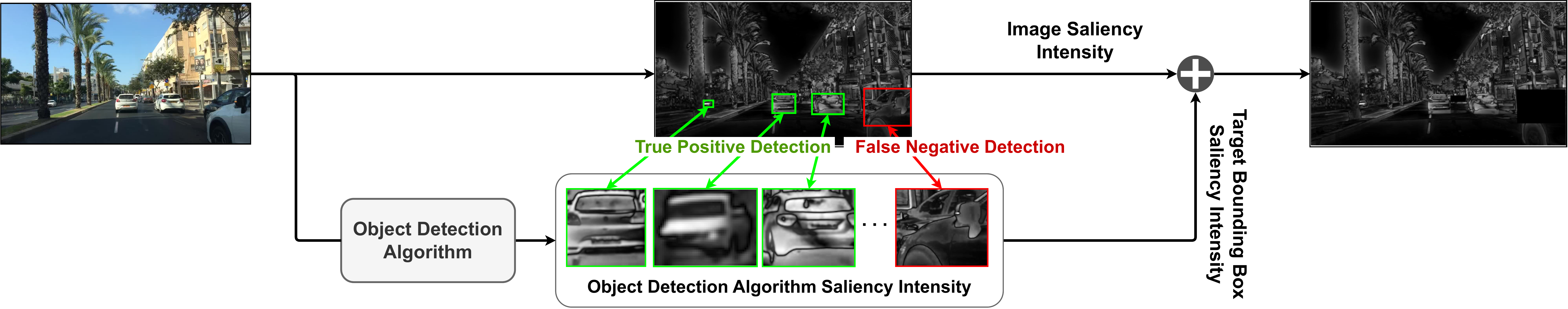}
\caption{Modified Multi-Object Fine-Grained Saliency Mapping Generation Method. The proposed object-based saliency map is a summation of the original saliency intensities and object detection-oriented saliency intensities}
\label{fig2:fine_grain_saliency}
\end{figure*}

\begin{figure}[h]
\centering
\includegraphics[width=8.5cm]{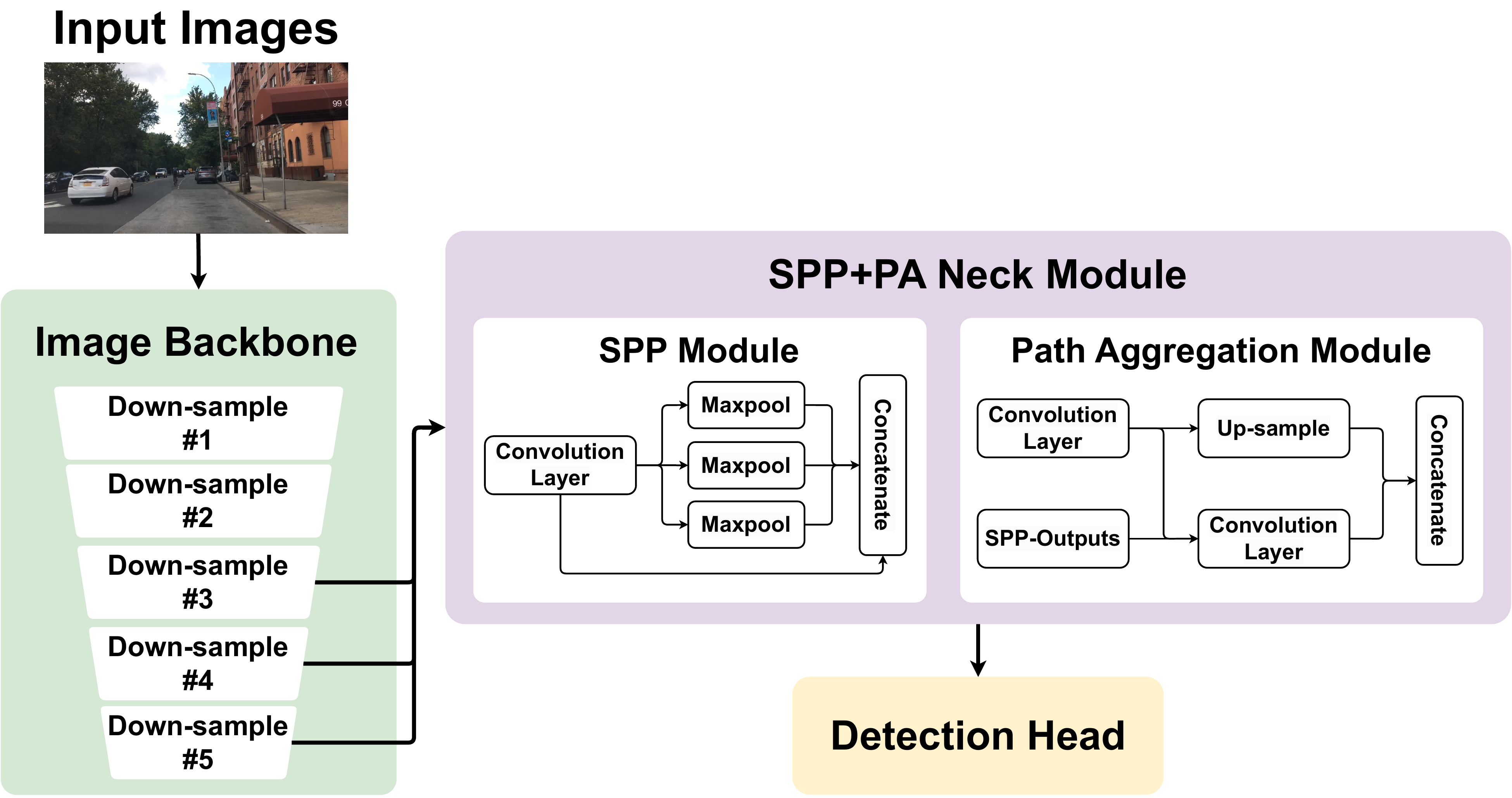}
\caption{YOLO-v4 Object Detection Model Architecture}
\label{fig3:yolo}
\end{figure}

\begin{figure}[h]
\centering
\includegraphics[width=7cm]{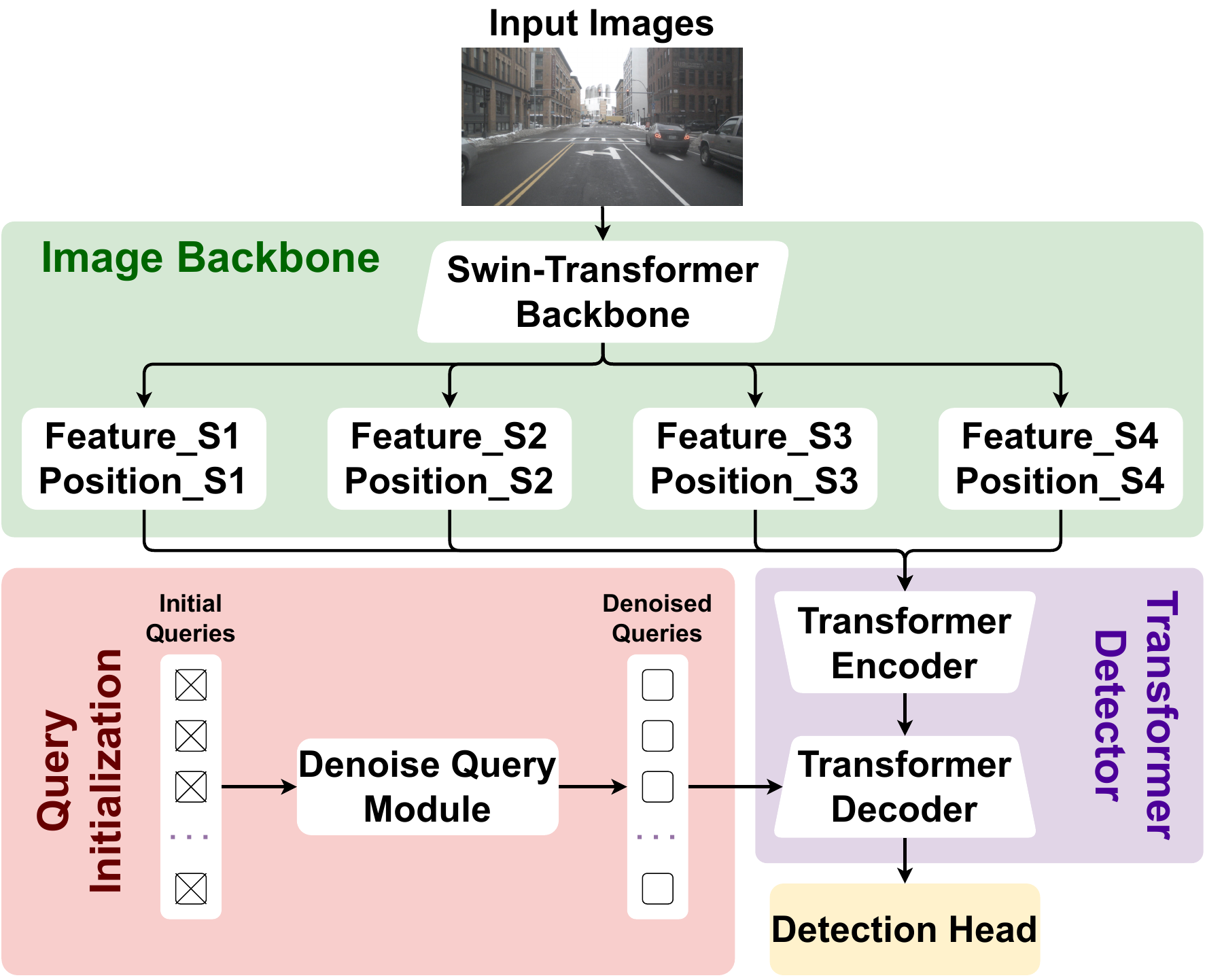}
\caption{Swin-Transformer-based DINO Detection Model Architecture}
\label{fig4:dino}
\end{figure}

\begin{figure*}[h]
\centering
\includegraphics[width=14.5cm]{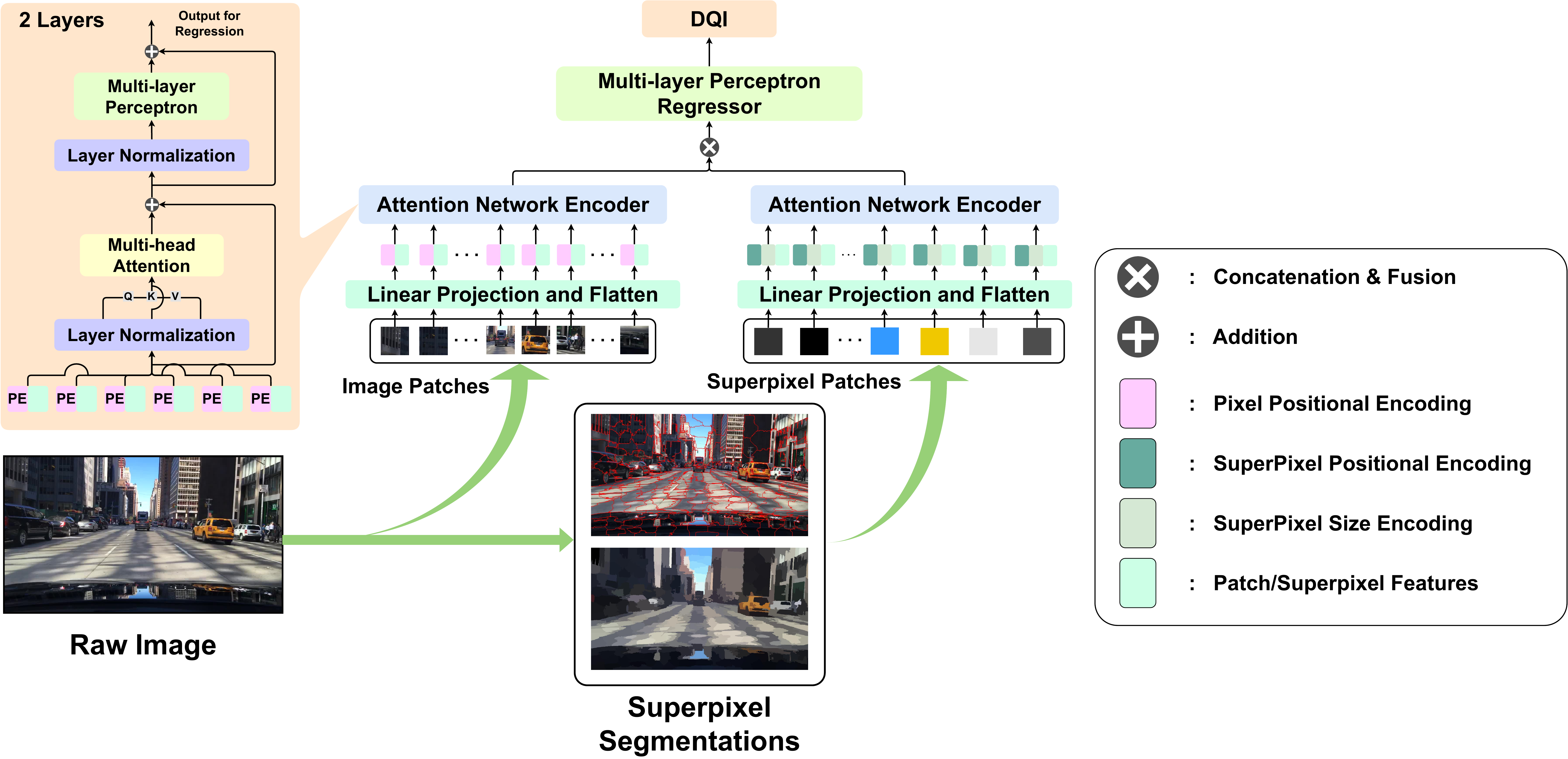}
\caption{SPA-NET Model Architecture. The proposed SPA-NET comprises superpixel and pixel module composed of attention-based networks}
\label{fig5:SPA-NET}
\end{figure*}

\subsection{DQI Generation}
The proposed DQI, as shown in Figure \ref{fig2:fine_grain_saliency}, is an evaluation metric to assess the object detection algorithm's detection quality. 
An object detection algorithm's detection accuracy is affected by two factors: (a) background image properties and (b) target object properties.
Background properties describe the overall image quality, such as brightness, blurriness, and contrast, and non-target object properties such as surrounding vegetation and buildings. 
Target object properties are the objects' quality, such as objects' sizes, colors, and their corresponding classes. 
To take both factors into account, we employ and modify fine-grained saliency map results as the DQI.
Although the original objective of fine-grained saliency is to identify the "most important object" within an image, our findings indicate that the fine-grained saliency map can effectively capture background objects as well. 
As a result, we have modified the fine-grained saliency map to include both the original fine-grained saliency intensities across the entire image and the saliency intensities specific to the detected objects based on the output of the object detection algorithm.

The proposed equation for the modified DQI is
\begin{equation}
I_{DQI} = mean(I_{SM}(x,y)+\sum_{i=1}^{k}c_i\cdot I_{obj}(x_m,y_m))
\label{eq:1}
\end{equation}
where \(I_{DQI}\) is the overall DQI, \(I_{SM}\) is the original fine-grained saliency mapping intensities, \(I_{obj}\) is the object saliency mapping intensities. 
In equation \ref{eq:1}, \(c_i\) is the object detection confidence score, and \(x\) and \(y\) are the width and height pixel positions, accordingly.
\subsubsection{Fine-Grained Saliency Mapping Generation}
To generate the image saliency map, we have employed the fine-grained method proposed by S. Montabone et. al \cite{MONTABONE2010391}. 
The core of the fine-grained method is the feature extraction module. 
Input images are then converted to greyscale images and transformed to different scales. According to \cite{MONTABONE2010391}, the fine-grained method obtain intensities by
\begin{equation}
I_{total}(x,y)=\sum_{i=1}^{n}{I_{sub_i}(x,y)}
\label{eq:2}
\end{equation}
where \(I_{sub_i}(x,y)\) is the \(i^{th}\) submap intensity and \(I_{total}(x,y)\) is the total map intensity. The submap intensity is computed by 
\begin{equation}
I_{sub}(x,y)=max\{c(x,y)-\sigma(x,y, \zeta), 0\}
\label{eq:3}
\end{equation}
where \(c(x,y)\) represents the center and $\sigma(x,y, \zeta)$ is the surroundings. The equations for obtaining \(c\) and $\sigma$ are
\begin{equation}
c(x,y)=i(x,y)
\label{eq:4}
\end{equation}
\begin{equation}
\sigma(x,y,\zeta)=\frac{rectSum((x-\zeta,y-\zeta,x+\zeta,y+\zeta)-i(x,y))}{(2\zeta+1)^2-1}
\label{eq:5}
\end{equation}
where \(rectSum\) is the rectangular sum, \(i(x,y)\) is the intensity of the corresponding pixel, and $\zeta$ is the filtering window size. 
In this study, both the whole image saliency mapping intensity and the object saliency mapping intensity are generated with the same method.

\subsubsection{Object Detection Algorithms}
To demonstrate the robustness of the proposed DQI, we evaluate its performance using two different algorithms: YOLO-v4 and DINO. 
YOLO-v4 utilizes a CNN-based architecture for both the backbone and detection processes, making it a real-time and dependable object detector. 
On the other hand, DINO employs a transformer-based architecture for the backbone and detection, resulting in a slower but more accurate object detector.
By utilizing these two distinct algorithms, we can assess the effectiveness of the DQI across different detection methodologies, encompassing both real-time efficiency and high accuracy. 
The architecture of YOLO-v4 and DINO are shown in Figure \ref{fig3:yolo} and \ref{fig4:dino} and the details about these two models are explained in \cite{Bochkovskiy2020YOLOv4OS}, \cite{zhang2021comparative}, and \cite{Zhang2022DINODW}. 
For the YOLO-v4, object detection bounding box threshold is 0.4 and 0.5 for the confidence score and intersection-over-union (IoU), respectively. 
For the DINO, the confidence score threshold is set to be 0.3.

\subsection{DQI Regression Model: SPA-NET}
Since DQI generation process requires ground-truth priors and the fine-grained saliency map generation is time-consuming, we propose a neural network model to regress the DQI score, which is named as SPA-NET, as shown in Figure \ref{fig5:SPA-NET}.
SPA-NET consists of two main modules: the pixel module and the superpixel module. 
The concept of combining superpixels and pixels is inspired by the approach used in 3D point cloud analysis, where detailed point features are combined with coarse voxel features \cite{9157234}. 
By leveraging this idea, we can extract both detailed local features from pixels and coarse but general features from superpixels.
In the pixel module, the raw input images undergo a transformation from a shape of [\(C\times H \times W\)] to [\(C\times512\times512\)].
Subsequently, they are fed into a pixel-based visual transformer (ViT) to extract relevant features \cite{Dosovitskiy2021AnII}.
For the superpixel module, the input images are initially segmented into superpixels. 
The extracted superpixel features are then passed through a superpixel-based visual transformer to capture informative features.
After implementing the pixel and superpixel modules, the extracted features are concatenated and fused together to facilitate the regression of the DQI score.
The subsequent sections of the document provide detailed explanations of both the pixel module and the superpixel module, outlining their respective functionalities and processes.

\subsubsection{pixel module}
The pixel module aims to extract local and detailed features from the image. 
The input image ([\(C\times512\times512\)]) are separated to patches with a size of \([3\times32\times32]\). 
Then, these patches are stacked and fed into attention modules for feature extraction. 
The attention module is a standard ViT attention module with eight attention heads and two attention layers. 
The output dimension from the attention module is 1024.

\subsubsection{superpixel module} 
The primary objective of the superpixel module is to extract coarse image features. 
It differs from the pixel module in terms of the preprocessing steps, input image shape, and positional encoding methods employed.
For image preprocessing, we utilize the fast Simple Linear Iterative Clustering (SLIC) method proposed by \cite{fast_slic} to generate superpixels. 
This method is reported to be 200\% faster than the traditional SLIC method. We set the number of superpixel segments to 500 in our implementation.
Regarding the input image, the data dimension shape is set as \([6\times500]\), where 500 represents the number of superpixels and 6 signifies the features associated with each superpixel. 
In this study, we choose the mean and standard deviation of the RGB channels as the input features for the superpixels.
For positional encoding, we incorporate two features: the size of the superpixel and its 2D position. 
The size of the superpixel is determined by the number of pixels within the bounded super-segments. 
The 2D positions correspond to the center point positions of the respective super-segments.
By incorporating these preprocessing steps, input image shape, and positional encoding methods, the superpixel module effectively captures coarse image features necessary for the subsequent fusion and regression processes.

After extracting the pixel and the superpixel features, the features are normalized and concatenated together for object detection quality regression. The regression module is a two-layer feedforward neural network where the hidden layer size is eighteen, and the output size is one.

\section{Datasets and Experiment}
We employ three open-source datasets, namely the BDD100K \cite{Yu2020BDD100KAD}, the KITTI \cite{Geiger2012CVPR}, and the nuScenes \cite{Caesar2020nuScenesAM}. 
The objective of applying three datasets is to (a) evaluate the correctness of the DQI approach and (b) prove the robustness of the proposed SPA-NET regression model. 
Besides the open-source datasets, we also conduct a real-world experiment by testing the SPA-NET on an autonomous vehicle.

\subsection{Open-source datasets}
\textbf{The BDD100K dataset}, released in 2018, stands as one of the largest naturalistic driving datasets available. 
This dataset comprises images and videos captured by a dash camera, along with corresponding inertial measurement unit (IMU) data. For our study, we utilize the image data and their corresponding annotations from this dataset to generate the DQI and develop the SPA-NET model.
Notably, BDD100K dataset offers extensive training and validation data, encompassing various driving environments and conditions such as different weather and times of the day. 
However, one drawback is that certain images in the dataset may contain artifacts and errors, such as camera misplacement or obstruction caused by windshield wipers.

\textbf{The KITTI dataset}, introduced in 2012, is specifically designed for autonomous vehicle perception. 
It employs LiDARs and stereo cameras to capture data on the surrounding environment during driving. 
In our paper, we use the fused image data along with corresponding annotations from this dataset. 
One notable advantage of the KITTI dataset is its high image quality, although a limitation is the relatively small sample size available.

\textbf{The nuScenes dataset}, developed by Motional in 2019, focuses on vehicle surrounding environment perception. 
It incorporates several LiDARs, radars, and cameras to gather comprehensive data. 
In our study, we analyze the images obtained from the multi-view cameras, along with their corresponding annotations. 
The nuScenes dataset offers benefits such as a large sample size and diverse driving conditions. 
However, it is worth mentioning that some images in the dataset may lack meaningful content due to the multi-view camera setup.
The implementation of the three open-source datasets is carried out independently in our study.
We have customized the separation of the training and validation samples, considering that we utilize the proposed DQI as the target label.

Initially, all three datasets included official sets of training samples, validation samples, and testing samples. 
However, since the annotations for the testing samples were not available, we solely focused on the training and validation samples for our analysis.
The official training samples from each dataset were used to train the object detection algorithms, specifically YOLO-v4 and DINO. These algorithms were trained using the respective datasets' training samples to optimize their performance.
Subsequently, the validation dataset was further divided into additional training samples and validation samples. 
This separation was done to facilitate the training and validation of the SPA-NET model. 
The training samples from the validation dataset were used to train the SPA-NET model, while the validation samples were employed to evaluate its performance.
Specific details regarding the split between the training and validation samples for all experiments can be found in Table \ref{table1:dataset}. 

\subsection{Real World Experiment}
\begin{figure}[h]
\centering
\includegraphics[width=7cm]{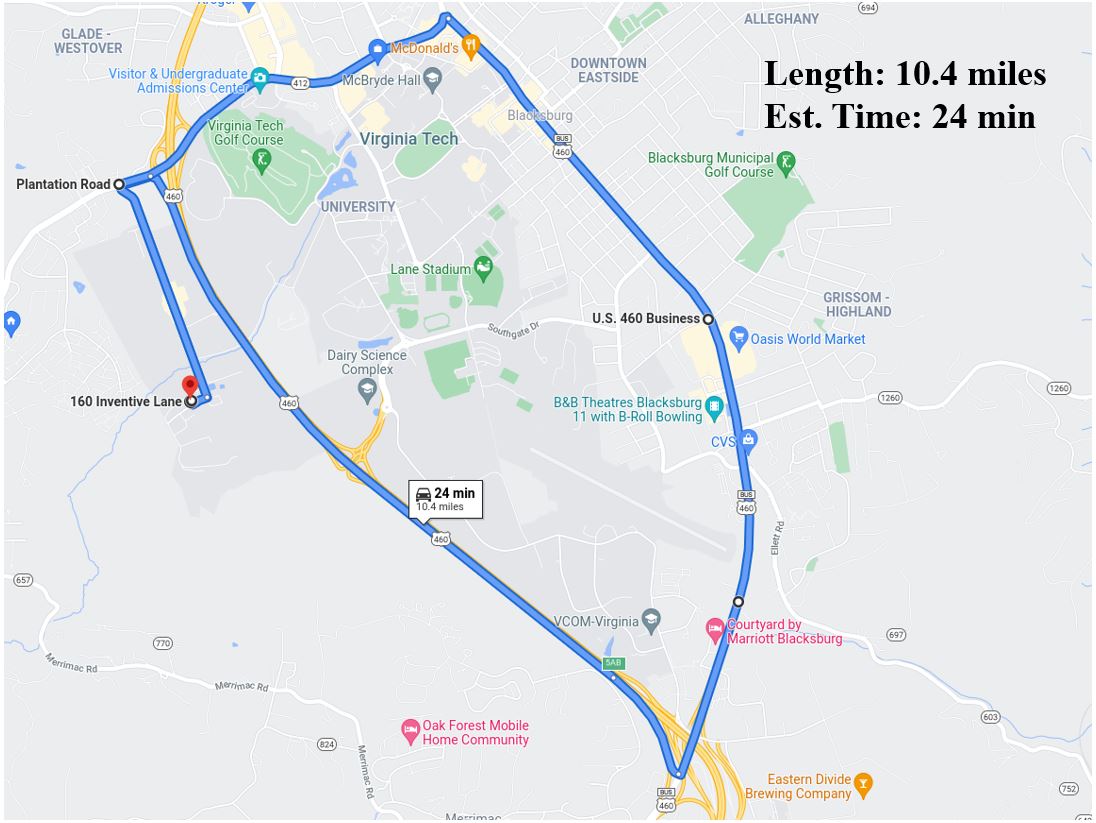}
\caption{Real World Experiment Driving Route and Estimated Time}
\label{fig6:route}
\end{figure}

\begin{figure}[h]
\centering
\includegraphics[width=7cm]{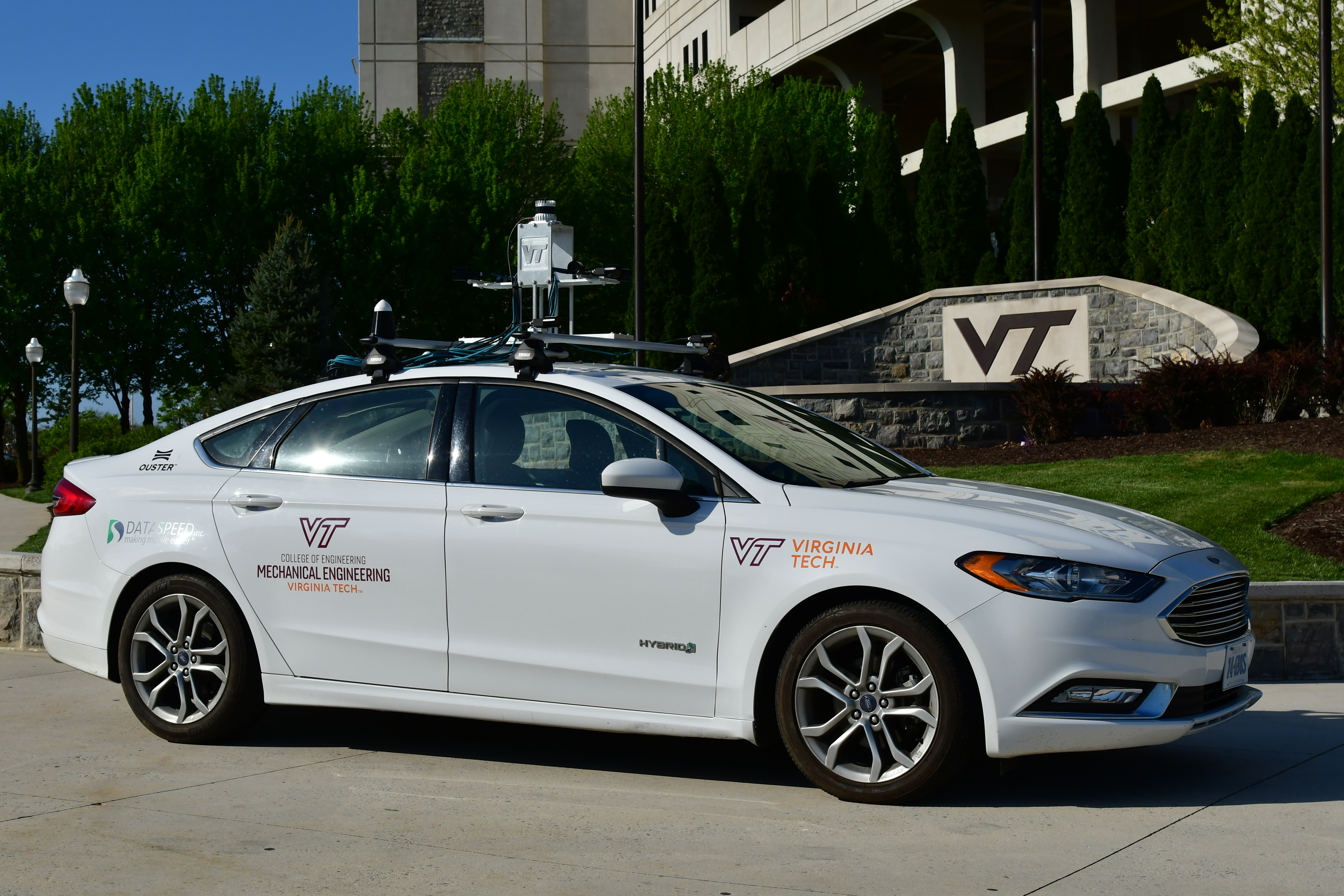}
\caption{Autonomous Vehicle for the Customized Experiment}
\label{fig7:vehicle}
\end{figure}

\begin{figure}[h]
\centering
\includegraphics[width=8.7cm]{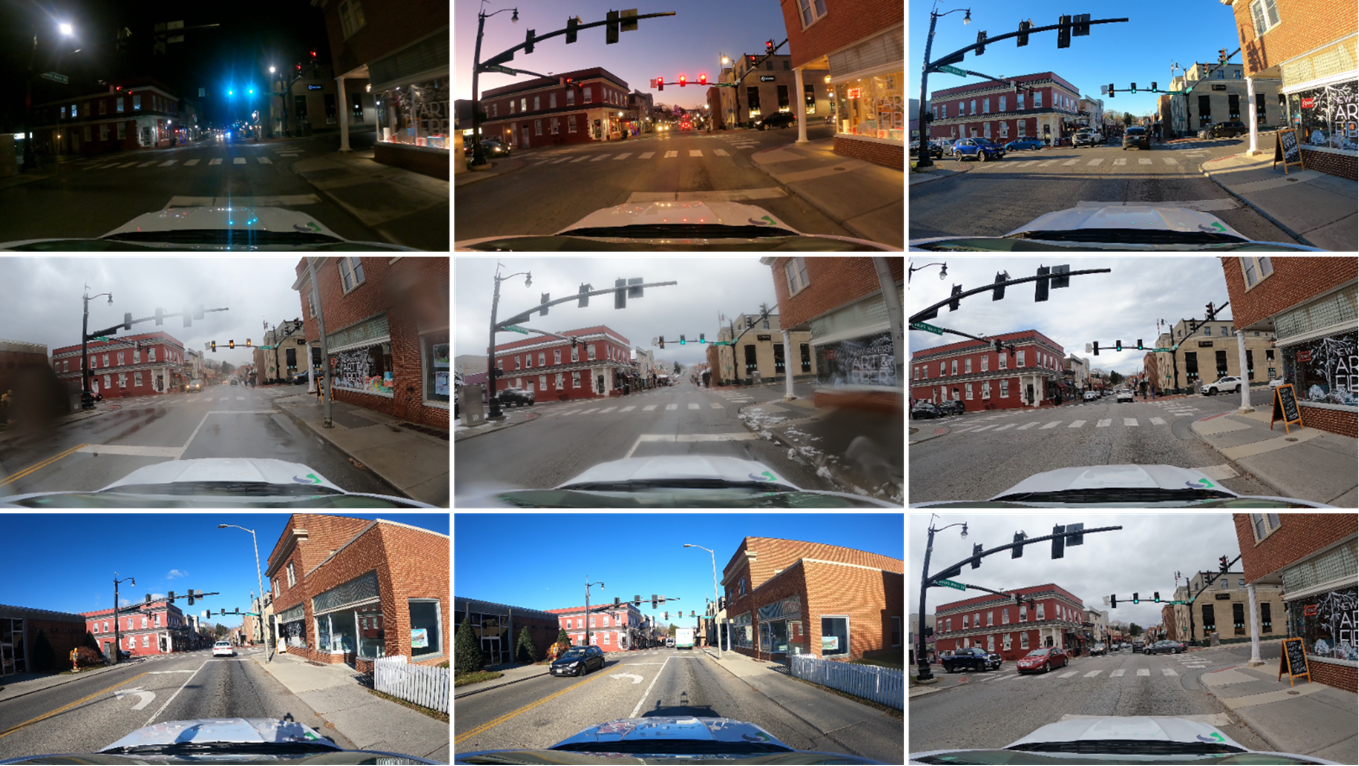}
\caption{Experiment Images Results at Similar Location with Different Weather and Time Conditions}
\label{fig8:time_weather}
\end{figure}

The objective of the experiment is to demonstrate the implementation of the proposed SPA-NET and qualitatively assess its performance under various real-world driving environments.
To achieve this, we record videos from the vehicle's front view and implemented the proposed SPA-NET to the images extracted from videos. 
The autonomous vehicle and the driving routes are shown in Figures \ref{fig6:route} and \ref{fig7:vehicle}. 
During testing, the driving environment contained morning, afternoon, evening, and night with different weather conditions such as cloudy, sunny, and rainy, as shown in Figure \ref{fig8:time_weather}.
By conducting the experiment in these real-world driving environments, we aim to evaluate the performance of SPA-NET in a practical setting. 
This assessment provides valuable insights into the model's ability to effectively assess object detection quality across different driving scenarios and environmental conditions.

\begin{table}[h]

\caption{Dataset Construction for Object Detection and Detection Quality}
\centering
\begin{tabular}{c|c|c|c|c}

\toprule
\multicolumn{5}{c}{\textbf{BDD100K}}                                                       \\\midrule
\textbf{Tasks}     & \textbf{Train} & \textbf{Validation} & \textbf{Test} & \textbf{Total} \\\midrule
Object Detection   & 70,000         & 10,000              & 20,000        & 100,000        \\\midrule
Detection Quality & 6,000          & 4,000               & N/A           & 10,000         \\\toprule
\multicolumn{5}{c}{\textbf{KITTI}}                                                         \\\midrule
\textbf{Tasks}     & \textbf{Train} & \textbf{Validation} & \textbf{Test} & \textbf{Total} \\\midrule
Object Detection   & 2,995          & 4,486               & 7,518         & 14,999         \\\midrule
Detection Quality & 2,692          & 1,794               & N/A           & 4,486          \\\toprule
\multicolumn{5}{c}{\textbf{nuScenes}}                                                      \\\midrule
\textbf{Tasks}     & \textbf{Train} & \textbf{Validation} & \textbf{Test} & \textbf{Total} \\\midrule
Object Detection   & 67,279         & 16,445              & 9,752         & 93,476         \\\midrule
Detection Quality & 9,867          & 6,578               & N/A           & 16,445         \\
\bottomrule
\end{tabular}
\label{table1:dataset}
\end{table}
\section{Results and Discussions}
The results section contains the DQI generation results among three open-source datasets, the proposed SPA-NET regression results, and the demo from the autonomous vehicle experiment.
\subsection{DQI Generation Results}
The DQI results are evaluated both qualitatively and quantitatively.
\subsubsection{Qualitatively Evaluation}
\begin{figure}[t]
\centering
\includegraphics[width=8.5cm]{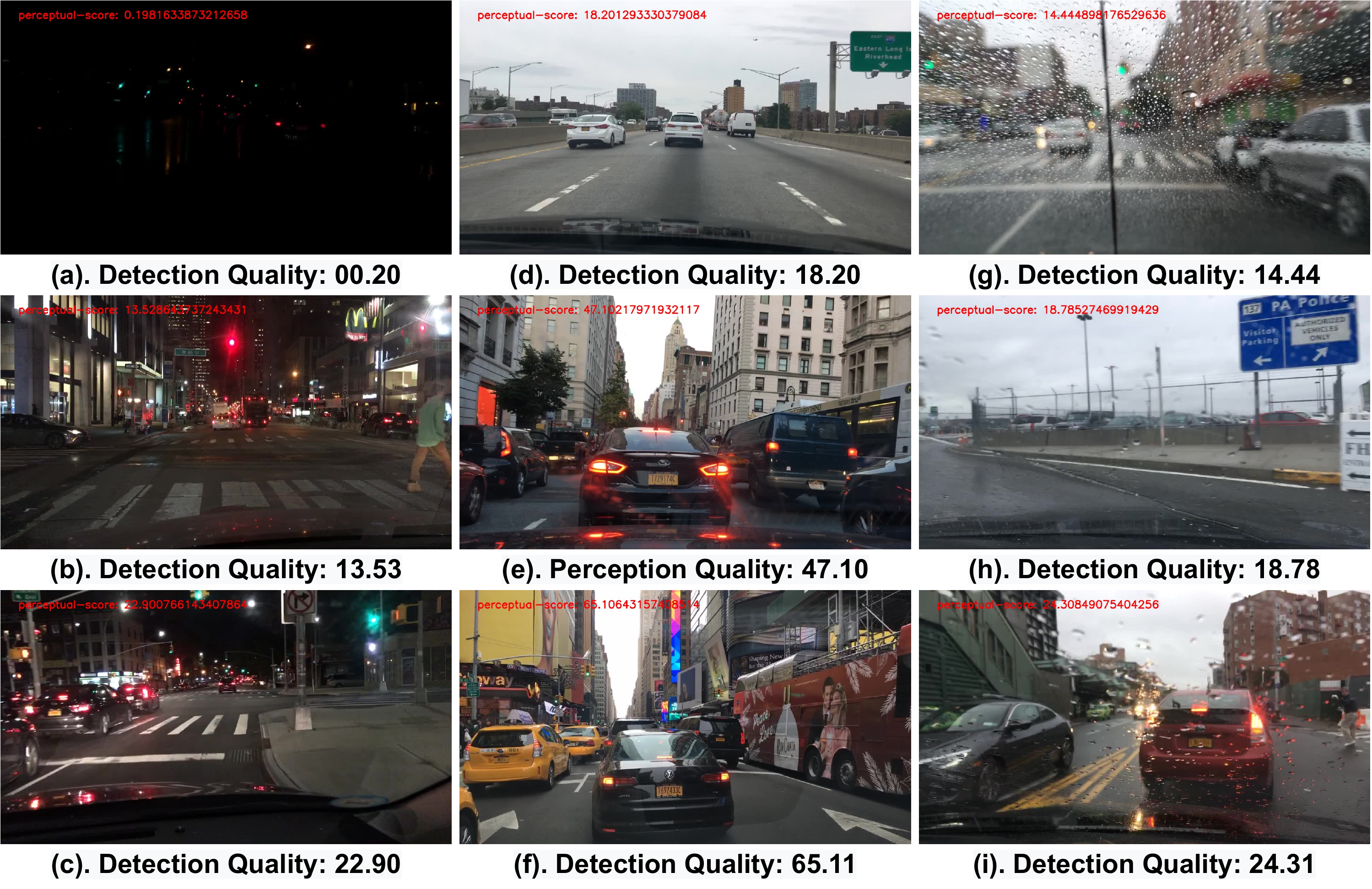}
\caption{BDD100K Dataset DQI Qualitative Observation}
\label{fig9:bdd_observe}
\end{figure}

\begin{figure}[t]
\centering
\includegraphics[width=8.5cm]{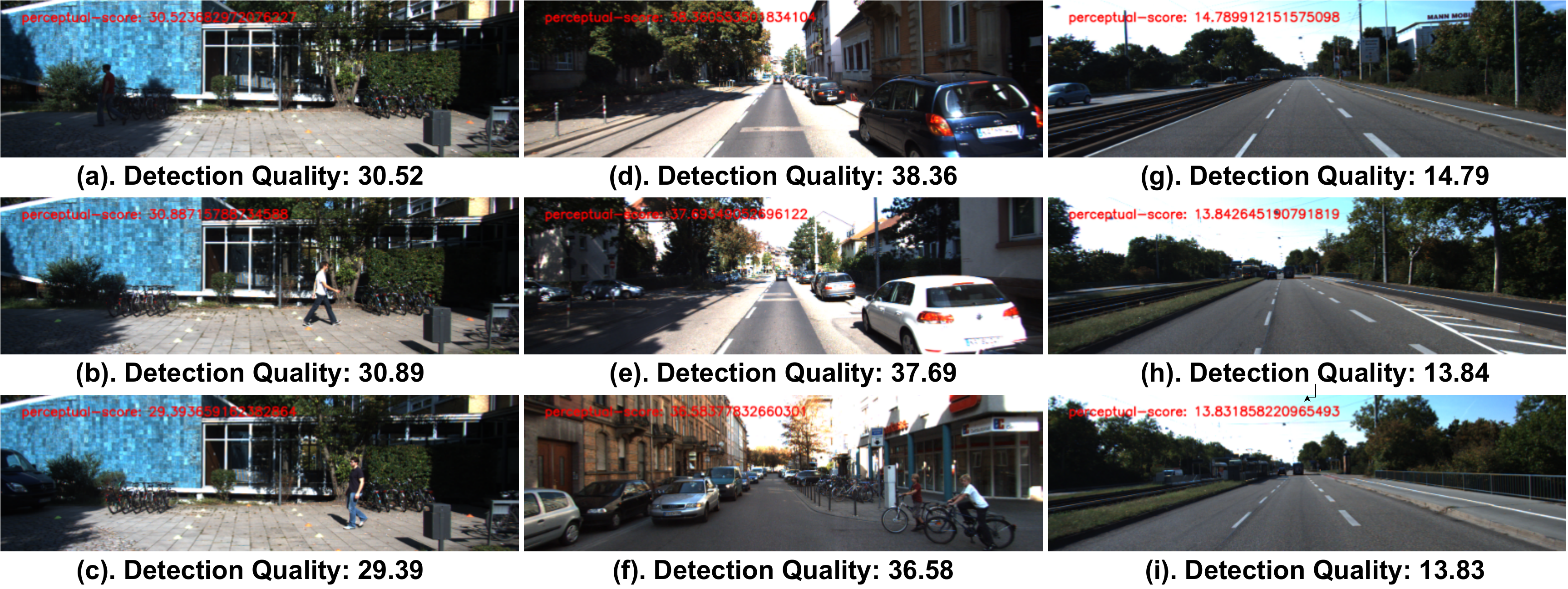}
\caption{KITTI Dataset DQI Qualitative Observation}
\label{fig10:kitti_observe}
\end{figure}

\begin{figure}[t]
\centering
\includegraphics[width=8.5cm]{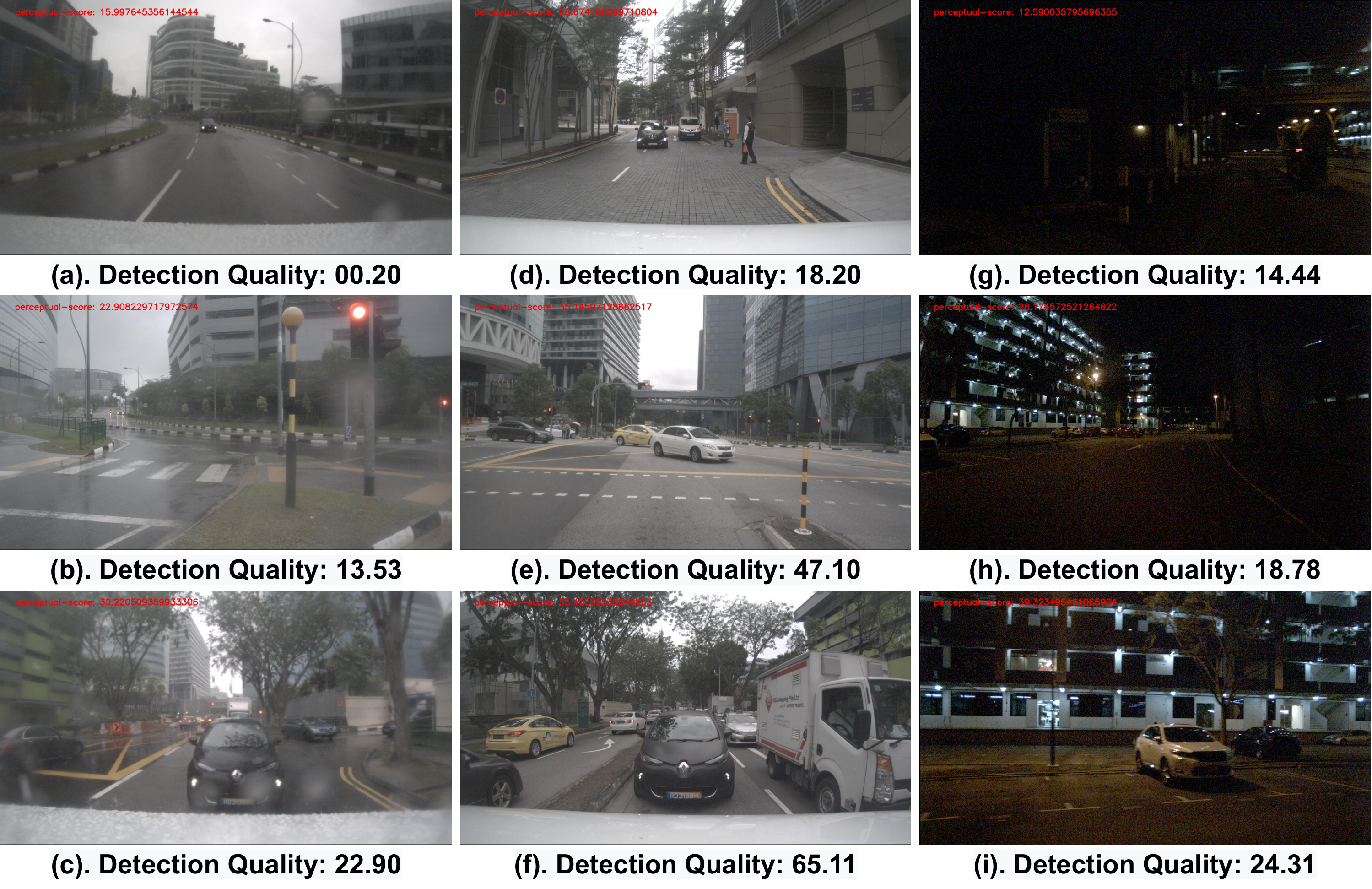}
\caption{nuScenes Dataset DQI Qualitative Observation}
\label{fig11:nuscene_observe}
\end{figure}

From Figures \ref{fig9:bdd_observe} to \ref{fig11:nuscene_observe}, we can observe that both the properties of the background and the target objects have an impact on the evaluation results of the detection quality. 
Moreover, it is evident that images with similar content tend to have similar detection quality indices (Figure \ref{fig10:kitti_observe}), which demonstrates the correctness of the generated detection quality index.
Regarding the background properties, examples in Figure \ref{fig10:kitti_observe} (a-c) indicate that increasing the darkness of the image background leads to a decrease in the detection quality index. 
This suggests that camera-based object detection algorithms may experience reduced detection quality under low-light conditions.
Examining the object properties in Figure \ref{fig10:kitti_observe} (d-f), we can observe that the DQI tends to be higher when the target object sizes are larger. 
This implies that larger objects are generally easier to detect, resulting in higher detection quality scores.
Additionally, Figure \ref{fig11:nuscene_observe} (a-c) and (d-e) demonstrate that images with similar content have similar perceptual scores. 
This further reinforces the validity of the proposed DQI generation method, as qualitative observations across different datasets and driving scenarios consistently align with our expectations.
Based on these qualitative and intuitive observations, we can conclude that the proposed DQI generation method is correct and valid. 
It effectively captures the impact of background and object properties on detection quality, providing reliable and meaningful quality scores.

\subsubsection{Quantitatively Evaluation}
The quantitative evaluation method contains two sections: (a) DQI score distribution analysis among all datasets, and (b) DQI average score analysis by adding synthetic artifacts to images among all datasets.

\begin{figure}[t]
\centering
\includegraphics[width=6.5cm]{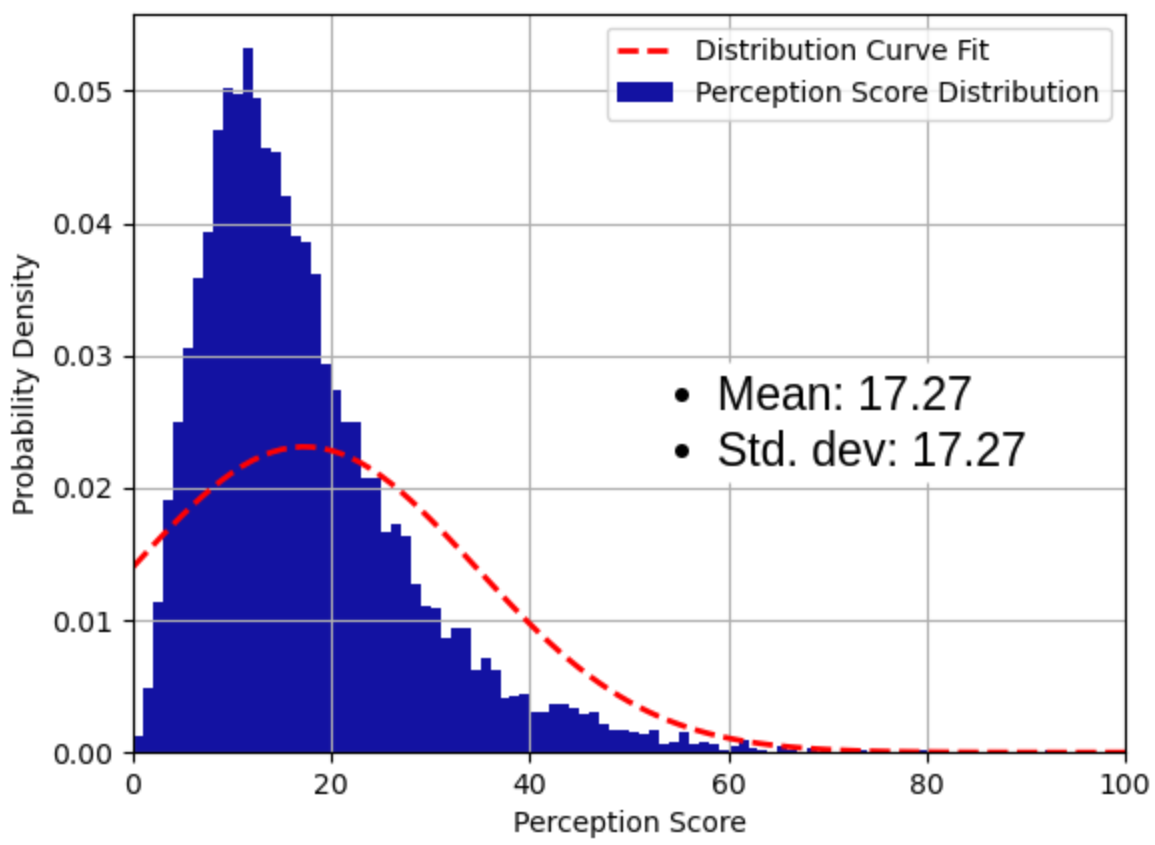}
\caption{BDD100K Dataset DQI Score Distribution Statistics}
\label{fig12:bdd_statistics}
\end{figure}

\begin{figure}[t]
\centering
\includegraphics[width=6.5cm]{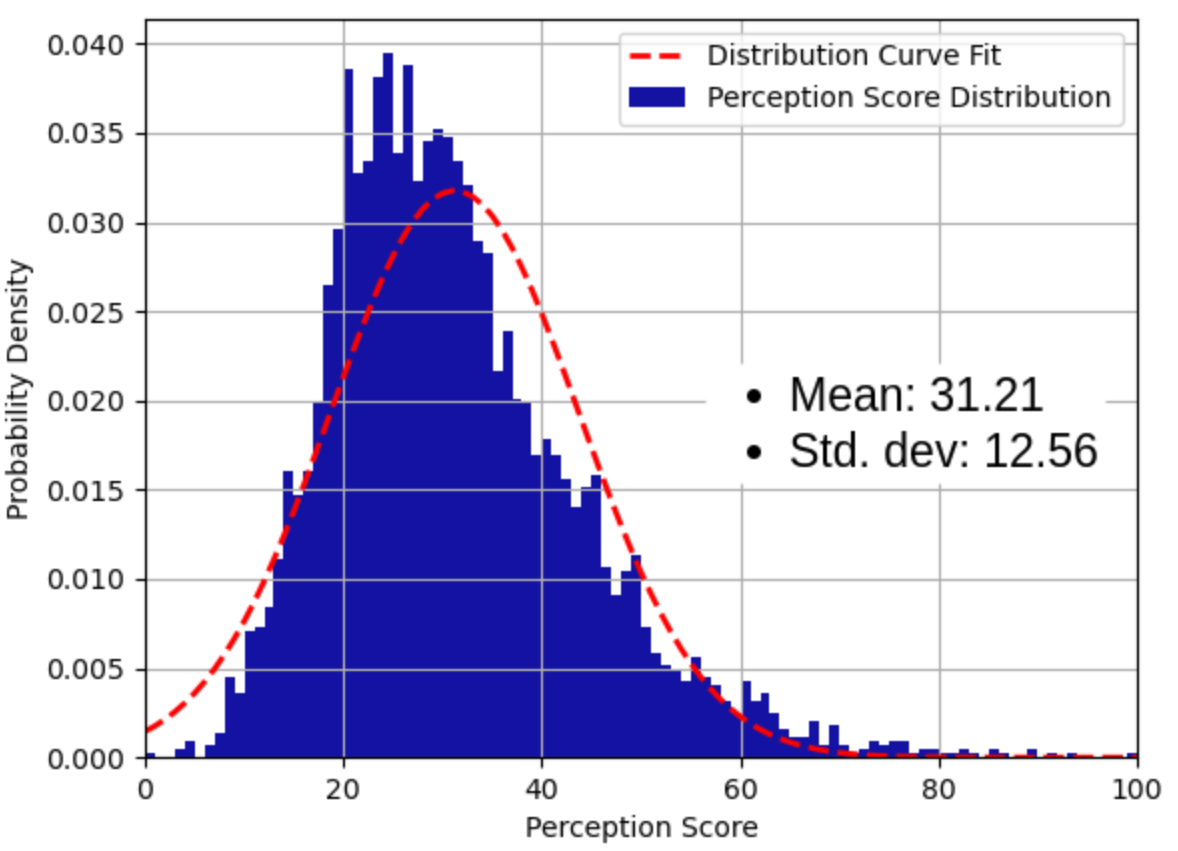}
\caption{KITTI Dataset DQI Score Distribution Statistics}
\label{fig13:kitti_statistics}
\end{figure}

\begin{figure}[t]
\centering
\includegraphics[width=6.5cm]{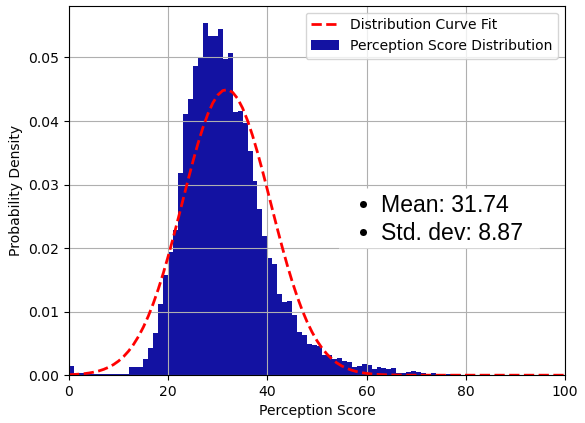}
\caption{nuScenes Dataset DQI Score Distribution Statistics}
\label{fig14:nuscene_statistics}
\end{figure}

\textbf{DQI Score Distribution Evaluation:}
Figure \ref{fig12:bdd_statistics}-\ref{fig14:nuscene_statistics} display the DQI score distribution among the BDD100K, KITTI, and nuScenes datasets.
The average DQI score for the BDD100K dataset is 17.27, which is lower than the nuScenes and KITTI datasets. 
This discrepancy can be attributed to the fact that the BDD100K dataset consists of images captured by in-vehicle dash cameras, which generally exhibit poorer visibility and lower image quality compared to out-vehicle fish-eye cameras. 
Furthermore, the BDD100K dataset is a naturalistic driving dataset collected from Uber drivers, resulting in a wider range of road conditions, times of the day, and weather variations compared to the other datasets.
In contrast, the KITTI and nuScenes datasets exhibit similar mean DQI scores, but the nuScenes dataset demonstrates a lower standard deviation. 
This is primarily because the image samples in the nuScenes dataset are captured along the same route, leading to greater similarity in the contents and colors among the entire dataset. 
The lower standard deviation suggests that regressing the DQI score for the nuScenes dataset is more challenging compared to the KITTI and BDD100K datasets, as there is less variation to capture within the dataset.

\textbf{Synthetic Artifact Effects Evaluation:}
To further validate the correctness of the DQI generation results, we conduct an analysis by introducing synthetic artifacts to the images and comparing the resulting score differences. 
This approach allows us to study the impact of controlled artifact levels on the trending behavior of the DQI scores.
Artifacts such as fog, blurriness, brightness, and darkness are known to have a significant influence on the performance of object detection algorithms. 
In our study, we specifically focus on these artifacts and generate them using the Automold program \cite{auto_mold}. 
The Automold program offers a range of synthetic image transformations, enabling us to simulate the effects of these artifacts with different levels.
Notably, the "speed" feature in the Automold program is utilized to introduce blurriness to the images, simulating the behavior associated with varying vehicle speeds.

\begin{figure}[t]
\centering
\includegraphics[width=7cm]{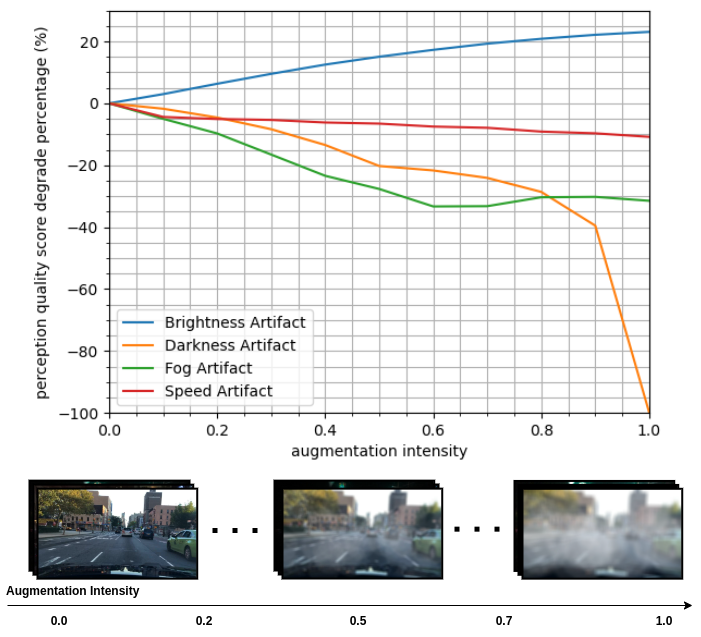}
\caption{BDD100K Dataset DQI Trend with Artificial Artifacts}
\label{fig15:bdd_trend}
\end{figure}

\begin{figure}[t]
\centering
\includegraphics[width=7cm]{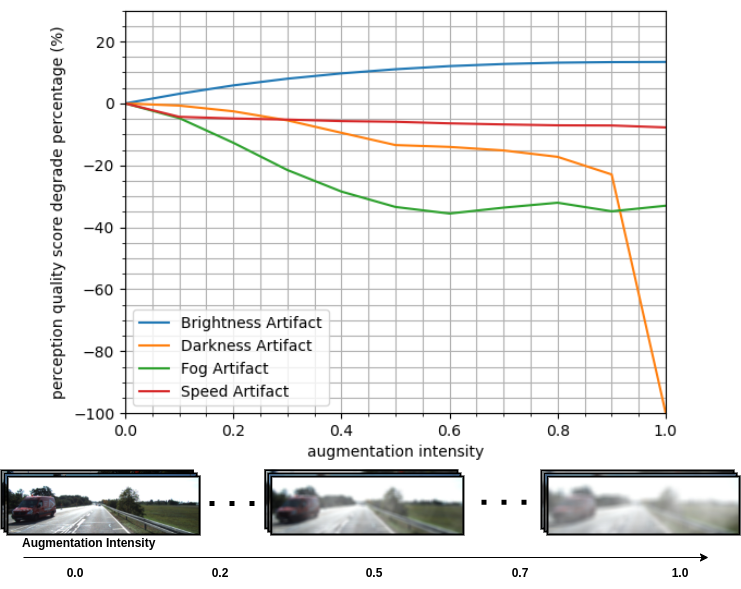}
\caption{KITTI Dataset DQI Trend with Artificial Artifacts}
\label{fig16:kitti_trend}
\end{figure}

\begin{figure}[t]
\centering
\includegraphics[width=7cm]{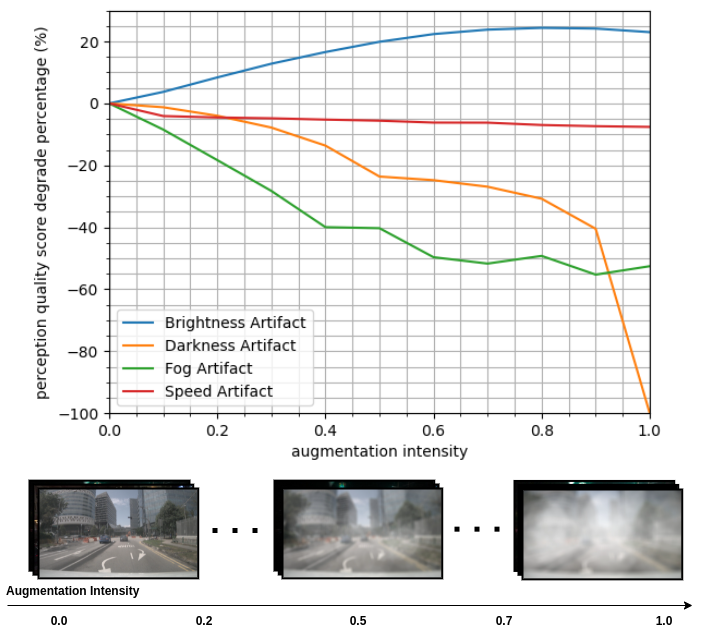}
\caption{nuScenes Dataset DQI Trend with Artificial Artifacts}
\label{fig17:nuscenes_trend}
\end{figure}

Figures \ref{fig15:bdd_trend} to \ref{fig17:nuscenes_trend} illustrate the scoring trends for brightness, darkness, fog, and speed artifacts across the BDD100K, KITTI, and nuScenes datasets. 
Based on these figures, several observations can be made:
(1) \textbf{Darkness and artificial fog:} The DQI score tends to decrease as the darkness and artificial fog intensity increase. This is consistent across all datasets. 
Darkness and fog are known to have a significant impact on object detection algorithm performance, making it more challenging for the algorithms to accurately detect objects in low-light or foggy conditions.
Therefore, lower DQI scores with more severe artifacts agree with out expectations.
(2) \textbf{Blurriness:} The DQI score shows a slight decrease as the blurriness intensity increases. 
This suggests that higher levels of blurriness negatively affect object detection performance, causing a decrease in the DQI score. 
However, the decrease in score is relatively small compared to darkness and fog, indicating that blurriness has a relatively milder impact on detection quality.
This trend is also agreed with our expectation, as blurriness affects the object detection algorithms' performance but not as much as the fog and darkness factors.
(3) \textbf{Brightness:} Increasing the image brightness results in an increase in the overall DQI score. 
Across all datasets, adding the brightness artifact leads to a maximum increase of 20\% in the DQI score. 
This observation aligns with the fact that saliency mapping intensity is influenced by image brightness. 
As the brightness of an image improves, the corresponding saliency intensity also improves, resulting in a higher DQI score.
It is important to note that in the Automold setup, a brightness factor of 1 does not necessarily mean the image turns completely white. 
The brightness factor in the Automold program operates on a specific scale, and even for large brightness factors, the image does not reach a pure white appearance. 
This explains why the DQI score can continue to increase even with larger brightness factors, as there is still room for further improvement in saliency mapping intensity.

\subsection{SPA-NET}
The following section presents the results of the SPA-NET model on multiple open-source datasets, as well as ablation studies conducted on the individual modules.
\subsubsection{Model Performance on Open-source Dataset}
In the DQI task, which shares similarities with the B-IQA task, we conduct an experiment where we implement several state-of-the-art B-IQA regression networks \cite{8576582, 9506075, 9156687} on the DQI task and compare the results with the proposed SPA-NET.
All models are trained and validated on NVIDIA A100 80G GPU with an Adam optimization algorithm. 
The number of the epoch is set to be 50, and the learning rate is a sinusoidal decay that ranges from 2e-5 to 1e-6.

The overall regression results are shown in Table \ref{table2:SPA-NET} and Table \ref{table3:SPA-NET_DINO}, and the evaluation metrics are Pearson's linear correlation coefficient (PLCC) \cite{doi:10.1098/rspl.1895.0041}, Spearman rank-order correlation coefficient (SRCC) \cite{10.5555/2531420}, and R-squared (R2) values \cite{GVK022791892}. 
Table \ref{table2:SPA-NET} presents the regression performance with YOLO-v4 as the detector and Table \ref{table3:SPA-NET_DINO} presents the regression performance with DINO as the detector. The equations for PLCC, SRCC, and R2 are
\begin{equation}
PLCC=\frac{\sum (predict)(target)}{\sqrt{\sum(predict)^2(target)^2}}
\end{equation}

\begin{equation}
predict = y_{predict}-mean(y_{predict})
\end{equation}

\begin{equation}
target = y_{target}-mean(y_{target})
\end{equation}
where \(y_{predict}\) is the predicted score, \(y_{target}\) is the ground truth score.

\begin{equation}
SRCC=\frac{cov(R(y_{predict}),R(y_{target}))}{\sigma_{R(y_{target})}\sigma_{R(y_{predict})}}
\end{equation}
where \(R(y_{predict}\) is the rank of the predicted score and \(R(y_{predict}\) is the rank of the ground truth.

\begin{equation}
R^2=1-\frac{RSS}{TSS}
\end{equation}
where \(RSS\) is the total sum of residual squares and \(TSS\) is the total sum of squares.

\textit{\textbf{BDD100K Dataset Results:}}  
For the BDD100K dataset, the proposed SPA-NET model's performance is the highest among all the other SOTA algorithms for R2 and the PLCC results, and only 0.2\% lower than the highest SRCC results. 
The BDD100K dataset training data sample size is 7,000, and the model is not pre-trained on any large image dataset such as the ImageNet. 
These prove that the proposed SPA-NET model can learn the detection quality features easily. 
Furthermore, the results indicate that the SPA-NET can be trained on small and local machines without complex and time-consuming pre-train steps.

\textit{\textbf{KITTI Dataset Results:}}  
For the KITTI dataset, the proposed SPA-NET achieves SOTA among the R2, the SRCC, and the PLCC results. Compared with the BDD100K dataset, the overall regression performance for all models on the KITTI dataset is lower. 
This is because of the small training sample size. 
According to Figure \ref{fig13:kitti_statistics}, even though the KITTI dataset's detection quality score is also well spread (standard deviation: 12.56), the training sample size is only around 2,600. 
Nevertheless, the SPA-NET performances are 26.45\%, 6.99\%, and 6.51\% higher than the second highest results on the R2, SRCC, and PLCC, which indicate that the SPA-NET can learn the image feature more efficiently and achieve better performance with a smaller dataset.

\textit{\textbf{nuScenes Dataset Results:}}  
For the nuScenes dataset, the SPA-NET is the 2nd highest for all evaluation metrics. 
The proposed model results are 1.6\%, 1.3\%, and 1.6\% lower than the Res50-Transformer model. 
According to Figure 14, even with 10,000 training samples, the nuScenes dataset standard deviation is 8.87, which is lower than the BDD100K and the KITTI datasets. 
This implies that the nuScenes dataset image perception features are more similar to each other compared with the other two datasets, which cause the model hard to obtain higher results. 
Even though the proposed SPA-NET performance is slightly lower than the Res50-Transformer model, the overall parameter size is smaller than the SOTA model, which leads to a lower computation load.

In Summary, our proposed SPA-NET achieves either the highest or the second highest results compared with other SOTA models. 
Moreover, the SPA-NET does not require the complex pre-training steps and extracts the image detection quality features more efficiently.

\begin{table}[]
\centering
\caption{SPA-NET Regression Results Among All Datasets for YOLO-v4 Detector}
\begin{tabular}{cccc}
\toprule
\rowcolor[HTML]{C0C0C0} 
\multicolumn{4}{c}{\cellcolor[HTML]{C0C0C0}
\textbf{BDD100K}}                                                                                                                \\ \midrule
                                                 & \textit{R2}                            & \textit{SRCC}                          & \textit{PLCC}                          \\ \midrule
\rowcolor[HTML]{C0C0C0} 
\textit{BiLinear \cite{8576582}}                                & 0.278                                  & 0.894                                  & 0.844                                  \\
\textit{Res50+Transformer \cite{9506075}}                       & 0.722                                  & 0.904                                  & 0.889                                  \\
\rowcolor[HTML]{C0C0C0} 
\textit{Res50+Hyper \cite{9156687}}                             & 0.748                                  & 0.897                                  & 0.882                                  \\
\textit{Vit-Baseline}                            & 0.649                                  & 0.893                                  & 0.881                                  \\
\rowcolor[HTML]{C0C0C0} 
{\color[HTML]{CB0000} \textit{\textbf{SPA-NET}}} & {\color[HTML]{CB0000} \textbf{0.777}}  & {\color[HTML]{CB0000} \textbf{0.902}}  & {\color[HTML]{CB0000} \textbf{0.891}}  \\ \toprule
\rowcolor[HTML]{EFEFEF} 
\multicolumn{4}{c}{\cellcolor[HTML]{EFEFEF}\textbf{KITTI}}                                                                                                                  \\ \midrule
\rowcolor[HTML]{C0C0C0} 
                                                 & \textit{R2}                            & \textit{SRCC}                          & \textit{PLCC}                          \\ \midrule
\textit{BiLinear \cite{8576582}}                                & NaN                                    & NaN                                    & NaN                                    \\
\rowcolor[HTML]{C0C0C0} 
\textit{Res50+Transformer \cite{9506075}}                       & 0.436                                  & 0.802                                  & 0.704                                  \\
\textit{Res50+Hyper \cite{9156687}}                             & 0.223                                  & 0.796                                  & 0.785                                  \\
\rowcolor[HTML]{C0C0C0} 
\textit{Vit-Baseline}                            & 0.545                                  & 0.838                                  & 0.833                                  \\
{\color[HTML]{CB0000} \textit{\textbf{SPA-NET}}} & {\color[HTML]{CB0000} \textbf{0.741}} & {\color[HTML]{CB0000} \textbf{0.901}} & {\color[HTML]{CB0000} \textbf{0.891}} \\ \toprule
\rowcolor[HTML]{C0C0C0} 
\multicolumn{4}{c}{\cellcolor[HTML]{C0C0C0}\textbf{NuScene}}                                                                                                                \\ \midrule
                                                 & \textit{R2}                            & \textit{SRCC}                          & \textit{PLCC}                          \\ \midrule
\rowcolor[HTML]{C0C0C0} 
\textit{BiLinear \cite{8576582}}                                & 0.265                                  & 0.834                                  & 0.707                                  \\
\textit{Res50+Transformer \cite{9506075}}                       & 0.534                                  & 0.831                                  & 0.813                                  \\
\rowcolor[HTML]{C0C0C0} 
\textit{Res50+Hyper \cite{9156687}}                             & 0.465                                  & 0.793                                  & 0.774                                  \\
\textit{Vit-Baseline}                            & 0.334                                  & 0.817                                  & 0.795                                  \\
\rowcolor[HTML]{C0C0C0} 
{\color[HTML]{CB0000} \textit{\textbf{SPA-NET}}} & {\color[HTML]{CB0000} \textbf{0.525}} & {\color[HTML]{CB0000} \textbf{0.820}} & {\color[HTML]{CB0000} \textbf{0.800}} \\ \bottomrule
\end{tabular}
\label{table2:SPA-NET}
\end{table}

\begin{table}[]
\centering
\caption{SPA-NET Regression Results Among All Datasets for DINO Detector}
\begin{tabular}{cccc}
\toprule
\rowcolor[HTML]{C0C0C0} 
\multicolumn{4}{c}{\cellcolor[HTML]{C0C0C0}\textbf{BDD100K}}                                                                                                                \\ \midrule
                                                 & \textit{R2}                            & \textit{SRCC}                          & \textit{PLCC}                          \\ \midrule
\textit{Vit-Baseline}                            & 0.720                                  & 0.887                                  & 0.874                                  \\
\rowcolor[HTML]{C0C0C0} 
{\color[HTML]{CB0000} \textit{\textbf{SPA-NET}}} & {\color[HTML]{CB0000} \textbf{0.759}}  & {\color[HTML]{CB0000} \textbf{0.904}}  & {\color[HTML]{CB0000} \textbf{0.897}}  \\ \midrule
\rowcolor[HTML]{EFEFEF} 
\multicolumn{4}{c}{\cellcolor[HTML]{EFEFEF}\textbf{KITTI}}                                                                                                                  \\ \midrule
\rowcolor[HTML]{C0C0C0} 
                                                 & \textit{R2}                            & \textit{SRCC}                          & \textit{PLCC}                          \\ \midrule
\rowcolor[HTML]{C0C0C0} 
\textit{Vit-Baseline}                            & 0.238                                  & 0.745                                  & 0.752                                  \\
{\color[HTML]{CB0000} \textit{\textbf{SPA-NET}}} & {\color[HTML]{CB0000} \textbf{0.759}} & {\color[HTML]{CB0000} \textbf{0.898}} & {\color[HTML]{CB0000} \textbf{0.906}} \\ \midrule
\rowcolor[HTML]{C0C0C0} 
\multicolumn{4}{c}{\cellcolor[HTML]{C0C0C0}\textbf{NuScene}}                                                                                                                \\ \midrule
                                                 & \textit{R2}                            & \textit{SRCC}                          & \textit{PLCC}                          \\ \midrule
\rowcolor[HTML]{C0C0C0} 
\textit{Vit-Baseline}                            & 0.214                                  & 0.758                                  & 0.721                                  \\
\rowcolor[HTML]{C0C0C0} 
{\color[HTML]{CB0000} \textit{\textbf{SPA-NET}}} & {\color[HTML]{CB0000} \textbf{0.514}} & {\color[HTML]{CB0000} \textbf{0.831}} & {\color[HTML]{CB0000} \textbf{0.805}} \\ \bottomrule
\end{tabular}
\label{table3:SPA-NET_DINO}
\end{table}

\subsubsection{Ablation Study}
The ablation study tests the impact of different superpixel sizes and superpixel module's positional encoding on the SPA-NET performance.

\textit{\textbf{Ablation Study on Superpixel Size: }} 
We have tested the superpixel size from 0 to 500 with a space of 100, as shown in Table \ref{table4:superpixel}. 
According to the results, when increasing the superpixel size, the SPA-NET's regression performance is slightly improved. 
We believe that this is because the larger superpixel sizes result in longer superpixel input sequences. 
When increasing the input superpixel sequences, the attention layer can extract higher dimension features, which gives the model more freedom to learn the hidden features. 
However, increasing the input sequences can cause a higher computation load due to the dot-product operation in the attention layer.

\begin{table}[]
\centering
\caption{SPA-NET Performance on Different Superpixel Sizes}
\begin{tabular}{cccc}
\toprule
\textbf{Superpixel Size} & \textbf{R2} & \textbf{PLCC} & \textbf{SRCC} \\ \midrule
0                        & 0.649       & 0.893         & 0.881         \\
100                      & 0.747       & 0.901         & 0.888         \\
200                      & 0.771       & 0.901         & 0.888         \\
300                      & 0.773       & 0.900         & 0.889         \\
400                      & 0.774       & 0.901         & 0.889         \\
500                      & 0.777       & 0.902         & 0.891         \\ \bottomrule
\end{tabular}
\label{table4:superpixel}
\end{table}

\textit{\textbf{Ablation Study on Superpixel Module Positional Encoding: }} 
The importance of superpixel size and positional encoding is to provide the input sequence order information for the attention layer \cite{luo2021stable, Vaswani2017AttentionIA}. 
Since attention layer in the ViT backbone module use multi-head self-attention, introducing input feature sequence order can improve the model learning capability. 
According to the experiment results, the superpixel module's position and size encoding is proved to be effective for the model performance. 
When adding the superpixel positional encoding and the superpixel size encoding, the performance is 6.43\%, 0.33\%, and 0.45\% higher on the R2, the PLCC, and the SRCC results, as shown in Table \ref{table5:position}. 
We find that the positional encoding is still important for the detection quality assessment task, but it is not as important as other popular vision tasks such as object detection, segmentation, or object tracking tasks. 
The reason is because the detection quality does not require too much image spatial information compared with the aforementioned vision related tasks.

\begin{table}[]
\centering
\caption{SPA-NET Performance on Positional Encoding}
\begin{tabular}{cccc}
\toprule
\textbf{Position Encoding} & \textbf{R2} & \textbf{PLCC} & \textbf{SRCC} \\ \midrule
No Pos-Encoding                        & 0.712       & 0.897         & 0.885         \\
With Pos-Encoding                      & 0.777       & 0.902         & 0.891         \\ \bottomrule
\end{tabular}
\label{table5:position}
\end{table}

\subsection{Open-Source \& Customize Dataset Demonstration}
\begin{figure}[h]
\centering
\includegraphics[width=6.5cm]{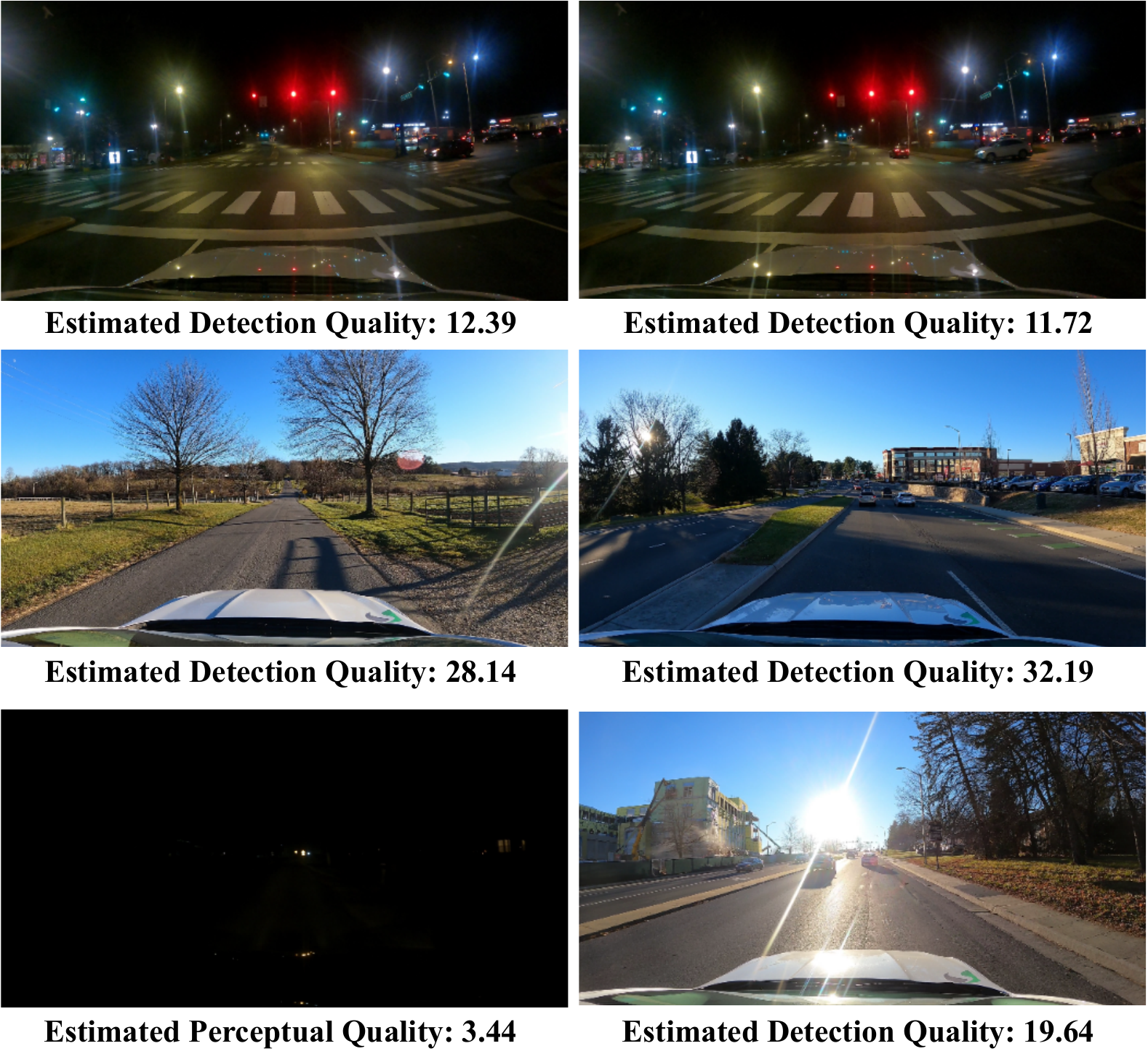}
\caption{Custom Dataset Detection Quality Index Results}
\label{fig18:Custom}
\end{figure}

\begin{figure}[h]
\centering
\includegraphics[width=7cm]{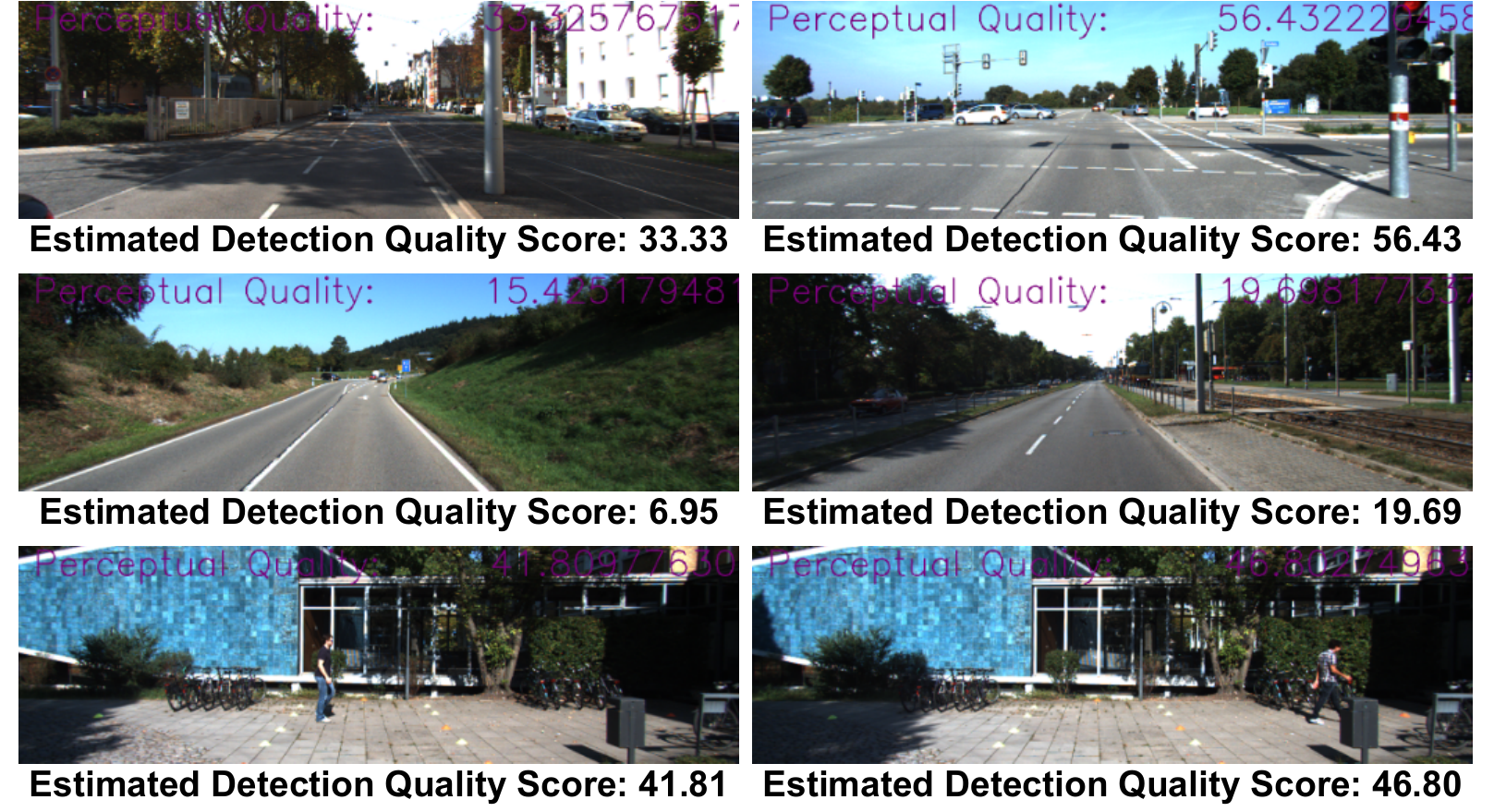}
\caption{KITTI Dataset Detection Quality Index Results Predicted by SPA-NET}
\label{fig19:KITTI}
\end{figure}

\begin{figure}[h]
\centering
\includegraphics[width=8.5cm]{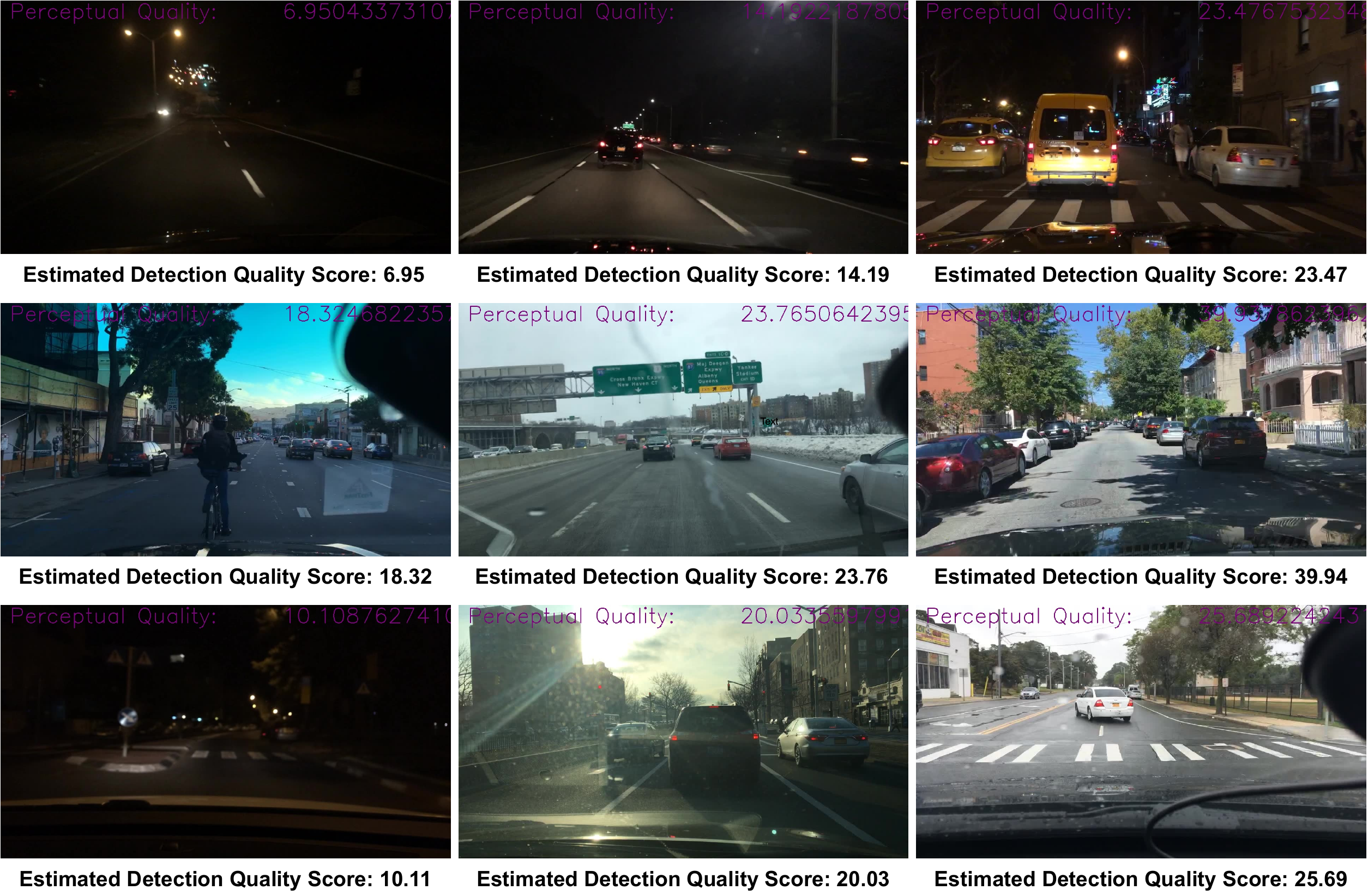}
\caption{BDD100K Dataset Detection Quality Index Results Predicted by SPA-NET}
\label{fig20:BDD}
\end{figure}

\begin{figure}[h]
\centering
\includegraphics[width=8.5cm]{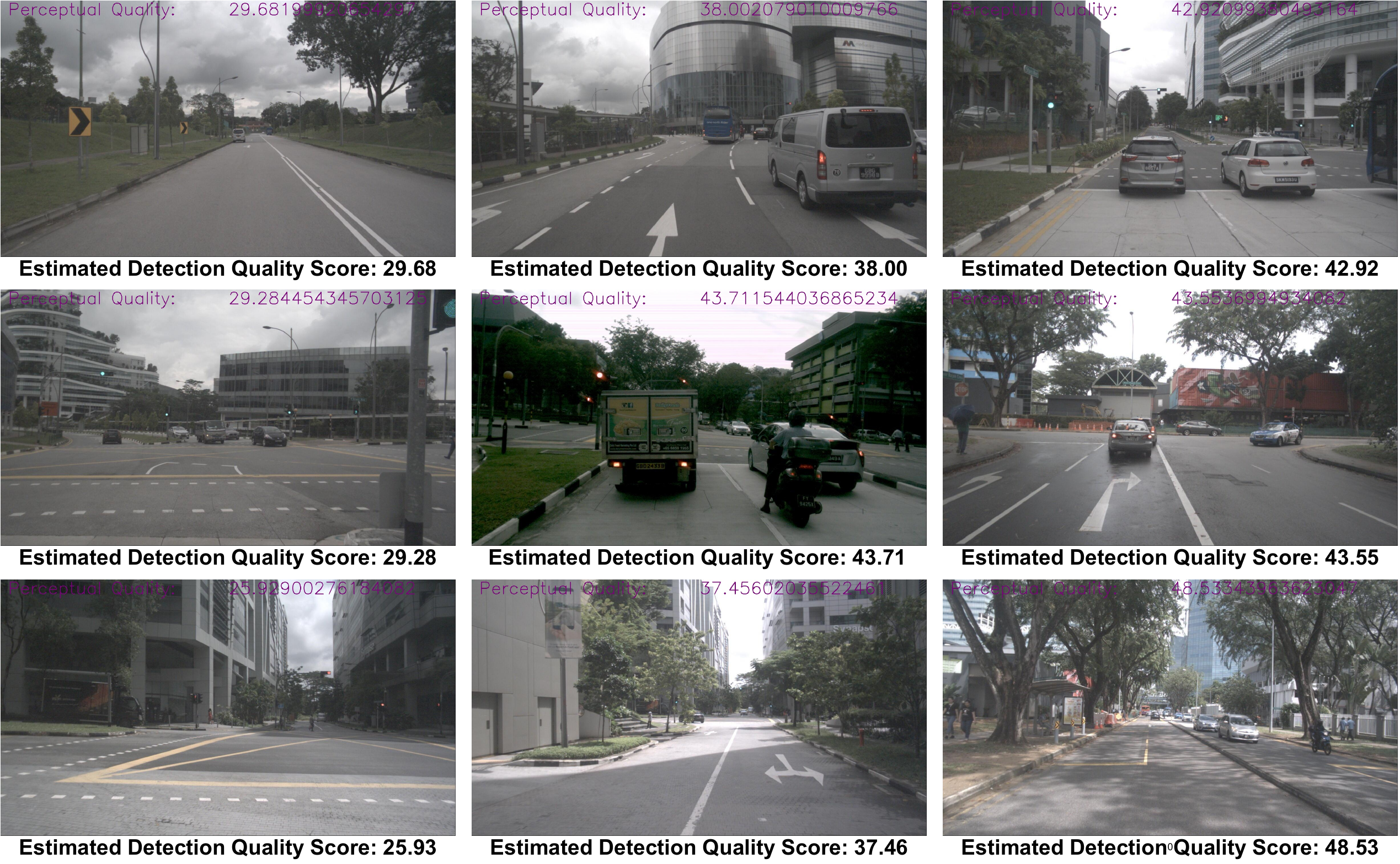}
\caption{nuScenes Dataset Detection Quality Index Results Predicted by SPA-NET}
\label{fig21:nuscenes}
\end{figure}

The objective of the open-source and customized dataset demonstration is to prove the proposed DQI robustness and correctness under real-world experiments. 
Furthermore, it can prove the performance of the proposed SPA-NET. 
Figure \ref{fig18:Custom} shows the selected customized dataset image samples and their corresponding DQI results. 
According to the figures, the proposed DQI evaluation metric is robust in a similar driving environment. 
Also, when the number of targets is increased and some objects are occluded, the general DQI score is dropped. 
Figure \ref{fig19:KITTI}, \ref{fig20:BDD}, and \ref{fig21:nuscenes} show the SPA-NET estimated detection quality score on the BDD100K, nuScenes, and KITTI datasets. 
According to KITTI's detection quality (Figure \ref{fig19:KITTI}), with similar image content and similar surrounding environment conditions, the SPA-NET predicted detection quality scores are similar. 
For the BDD100K's results (Figure \ref{fig20:BDD}), we concluded that the DQI score is strongly correlated with the image content. 
For instance, in Figure \ref{fig20:BDD}'s top row, all three images are at night conditions, while the DQI are higher when the surrounding objects are clearer and close to the ego vehicle. 
This behavior also agrees with the object detection algorithm's performance. 
Thus, based on these open-source datasets' results, the proposed DQI generation method can correctly present the detection quality for the surrounding environment, and the novel SPA-NET can successfully predict the DQI.

\section{Conclusions \& Future Works}

This research paper introduces a unique evaluation metric called DQI, which aims to assess the quality of camera perception in autonomous vehicles. Through both qualitative and quantitative analyses, it has been demonstrated that this proposed metric accurately evaluates the performance of camera-based detection algorithms. Additionally, we have developed a regression model called SPA-NET, which effectively predicts the DQI scores. The regression performance of SPA-NET achieves SOTA results, with R2, SRCC, and PLCC scores of 0.741, 0.903, and 0.905 respectively in the BDD100K dataset, 0.603, 0.836, and 0.827 in the KITTI dataset, and 0.525, 0.820, and 0.800 in the nuScenes dataset.

While this work demonstrates promising results, there are still areas that can be improved upon. Future research endeavors could focus on extending the evaluation metric DQI to include video-based assessment, as object tracking plays a crucial role in autonomous vehicle perception. Additionally, further enhancements to the proposed SPA-NET model are needed, as it currently requires high computational resources and is unable to achieve real-time performance in embedded systems. We anticipate that the introduction of DQI will serve as an inspiration for other researchers who are interested in exploring the quality and reliability of perception sensors in autonomous vehicles.


\bibliography{literature_review}

\begin{IEEEbiography}[{\includegraphics[width=1in,height=1.25in,clip,keepaspectratio]{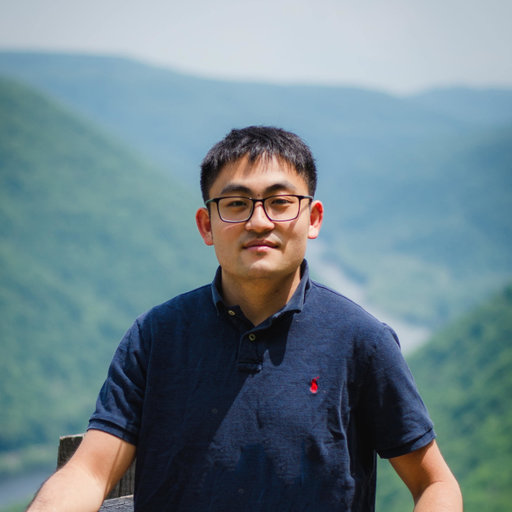}}]{Ce Zhang}
is with the Department of Mechanical Engineering at Virginia Tech. He received both his Bachelor's degree and Ph.D degree at Virginia Tech in 2018 and 2023, accordingly. Ce's research area is focused on autonomous vehicle perception, deep learning, and signal processing.
\end{IEEEbiography}

\begin{IEEEbiography}[{\includegraphics[width=1in,height=1.25in,clip,keepaspectratio]{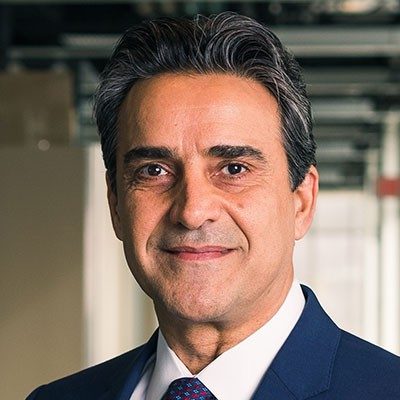}}]{Azim Eskandarian}
has been a Professor and Head of the Mechanical Engineering Department at Virginia Tech (VT) since August 2015. He became the Nicholas and Rebecca Des Champs chaired Professor in April 2018. He established the Autonomous Systems and Intelligent Machines laboratory at VT to conduct research in intelligent and autonomous vehicles, and mobile robotics. 

Prior to that, he was a Professor of Engineering and Applied Science at The George Washington University (GWU) and the founding Director of the Center for Intelligent Systems Research (1996-2015), the director of the Transportation Safety and Security University Area of Excellence (2002-2015), and the co-founder of the National Crash Analysis Center (1992) and its Director (1998-2002 \& 5/2013-7/2015). 
Earlier, he was an Assistant Professor at Pennsylvania State University, York, PA (1989-92) and worked as an engineer/project manager in industry (1983-89). 

He is the Editor-in-Chief of the IEEE Transactions on Intelligent Transportation Systems. He was awarded the IEEE ITS Society’s Outstanding Researcher Award in 2017 and the GWUs School of Engineering Outstanding Researcher Award in 2013. Dr. Eskandarian is a fellow of ASME, senior member of IEEE, and member of SAE professional societies. He received his BS, MS, and DSC degrees in Mechanical engineering from GWU, Virginia Tech, and GWU, respectively.
\end{IEEEbiography}

\end{document}